\documentclass[sigconf]{acmart}
\usepackage{fancyhdr}
\usepackage{amsfonts}
\usepackage{algorithmic}
\usepackage{textcomp}
\usepackage{xcolor}
\usepackage{siunitx}
\usepackage{subcaption}
\usepackage{multirow}
\usepackage{pifont}
\usepackage{algorithmic}
\usepackage{algorithm}
\usepackage{url}
\usepackage{bm}

\newif\ifmodify 
\modifytrue % CARE: comment this line to compile the final version
\ifmodify

\else

\fi
\newcommand{\M}{Quasar-ViT}

\newcommand{\rev}[1]{#1}

\AtBeginDocument{%
  }

\copyrightyear{2024}
\acmYear{2024}
\setcopyright{acmlicensed}\acmConference[ICS '24]{Proceedings of the 38th ACM International Conference on Supercomputing}{June 4--7, 2024}{Kyoto, Japan}
\acmBooktitle{Proceedings of the 38th ACM International Conference on Supercomputing (ICS '24), June 4--7, 2024, Kyoto, Japan}
\acmDOI{10.1145/3650200.3656622}
\acmISBN{979-8-4007-0610-3/24/06}

\begin{document}

%%
%% The "title" command has an optional parameter,
%% allowing the author to define a "short title" to be used in page headers.
\title{Quasar-ViT: Hardware-Oriented \underline{Qu}antization-Aware \underline{A}rchitecture \underline{S}e\underline{ar}ch for \underline{Vi}sion \underline{T}ransformers}

%%
%% The "author" command and its associated commands are used to define
%% the authors and their affiliations.
%% Of note is the shared affiliation of the first two authors, and the
%% "authornote" and "authornotemark" commands
%% used to denote shared contribution to the research.

\author{Zhengang Li\textsuperscript{1}, Alec Lu\textsuperscript{2}, Yanyue Xie\textsuperscript{1}, Zhenglun Kong\textsuperscript{1}, Mengshu Sun\text{$^\dagger$}, Hao Tang\textsuperscript{3}, Zhong Jia Xue\textsuperscript{2}, Peiyan Dong\textsuperscript{1}, Caiwen Ding\textsuperscript{4}, Yanzhi Wang\textsuperscript{1}, Xue Lin\textsuperscript{1}, Zhenman Fang\textsuperscript{2} \\
\textsuperscript{1}Northeastern University, 
\textsuperscript{2}Simon Fraser University, 
\textsuperscript{3}ETH Zurich, 
\textsuperscript{4}University of Connecticut \\
{E-mail:} \textsuperscript{1}\{li.zhen, xie.yany, kong.zhe, dong.pe, yanz.wang, xue.lin\}@northeastern.edu \\
\textsuperscript{2}\{alec\_lu, zjxue, zhenman\}@sfu.ca}
\thanks{$^\dagger$ MS is now afflicted with Beijing University of Technology; she contributed to this work during Ph.D at Northeastern University.}

\renewcommand{\shortauthors}{}

%%
%% The abstract is a short summary of the work to be presented in the
%% article.
\begin{abstract}
Vision transformers (ViTs) have demonstrated their superior accuracy for computer vision tasks compared to convolutional neural networks (CNNs). However, ViT models are often computation-intensive for efficient deployment on resource-limited edge devices.
This work proposes \M, a hardware-oriented quantization-aware architecture search framework for ViTs, to design efficient ViT models for hardware implementation while preserving the accuracy.
First, \M~trains a supernet using our row-wise flexible mixed-precision quantization scheme, mixed-precision weight entanglement, and supernet layer scaling techniques.
Then, it applies an efficient hardware-oriented search algorithm, integrated with hardware latency and resource modeling, to determine a series of optimal subnets from supernet under different inference latency targets. 
Finally, we propose a series of model-adaptive designs on the FPGA platform to support the architecture search and mitigate the gap between the theoretical computation reduction and the practical inference speedup. Our searched models achieve 101.5, 159.6, and 251.6 frames-per-second (FPS) inference speed on the AMD/Xilinx ZCU102 FPGA with 80.4\%, 78.6\%, and 74.9\% top-1 accuracy, respectively, for the ImageNet dataset, consistently outperforming prior works.
\end{abstract}

\maketitle

\section{Introduction} \label{sec:intro}

ViTs~\cite{raghu2021vision,dosovitskiy2020image, Touvron2021TrainingDI, yuan2021tokens} incorporate the attention mechanism~\cite{vaswani2017attention} to fulfill various computer vision tasks, by allowing all the pixels in an image to interact through transformer encoder blocks and thus achieving higher accuracy compared to CNNs. 
Table~\ref{tab:comparison_model} compares representative CNN and ViT models, i.e., ResNet~\cite{he2016deep}/ResNeXt~\cite{xie2017aggregated} and DeiT~\cite{Touvron2021TrainingDI} for the ImageNet dataset. 
DeiT-small (DeiT-S) with a comparable number of parameters and GMACs as ResNet-50 achieves even higher accuracy than ResNeXt-101, whose size is around 4$\times$ as that of DeiT-S.
DeiT-base (DeiT-B) with comparable size as ResNeXt-101 achieves 2.54\% higher top-1 accuracy.
\begin{table}[h]
\caption{Comparison of ResNets, ResNeXt, and DeiTs on ImageNet dataset. We choose DeiT without distilling token here, which represents state-of-the-art ViTs, as it can be directly trained on ImageNet-10k without pre-training on a massive dataset. }\label{tab:comparison_model}
\centering
% \tabcolsep 4pt
% \small
\resizebox{1.0 \columnwidth}{!}{\begin{tabular}{c|cccc }%{p{0.25\linewidth}p{0.25\linewidth}p{0.25\linewidth}p{0.25\linewidth}}
\toprule
Model & \#Params (M) & MACs (G) & Top-1 Acc. & Top-5 Acc. \\
\hline
ResNet-18 & 11.69 & 1.82 & 69.76\% & 89.08\% \\
ResNet-50 & 26.56 & 4.14 & 76.13\% & 92.86\% \\
ResNet-152 & 60.19 & 11.61 & 78.31\% & 94.05\% \\
ResNeXt-101 & 88.79 & 16.59 & 79.31\% & 94.53\% \\
\bf{DeiT-S} & 22.10 &  4.60 & 79.85\%  & 94.97\%  \\
\bf{DeiT-B} & 87.50 &  17.60  & 81.85\% & 95.59\%  \\
\bottomrule
\end{tabular}}
% \vspace{-0.4cm}
\end{table}

Despite ViTs' significant accuracy improvement, 
it is non-trivial to deploy ViT inference on resource-limited edge devices due to their huge model size and complex architectures. For example, even the lightweight ViT model DeiT-S~\cite{Touvron2021TrainingDI} has a model size of $22.10M$ parameters $\times$ $4 Bytes$ per floating-point parameter = $88.4MB$, presenting an overwhelming computing load and memory size for most edge devices. 

\begin{figure}[htb]
  \centering
  \includegraphics[width=1.0\columnwidth]{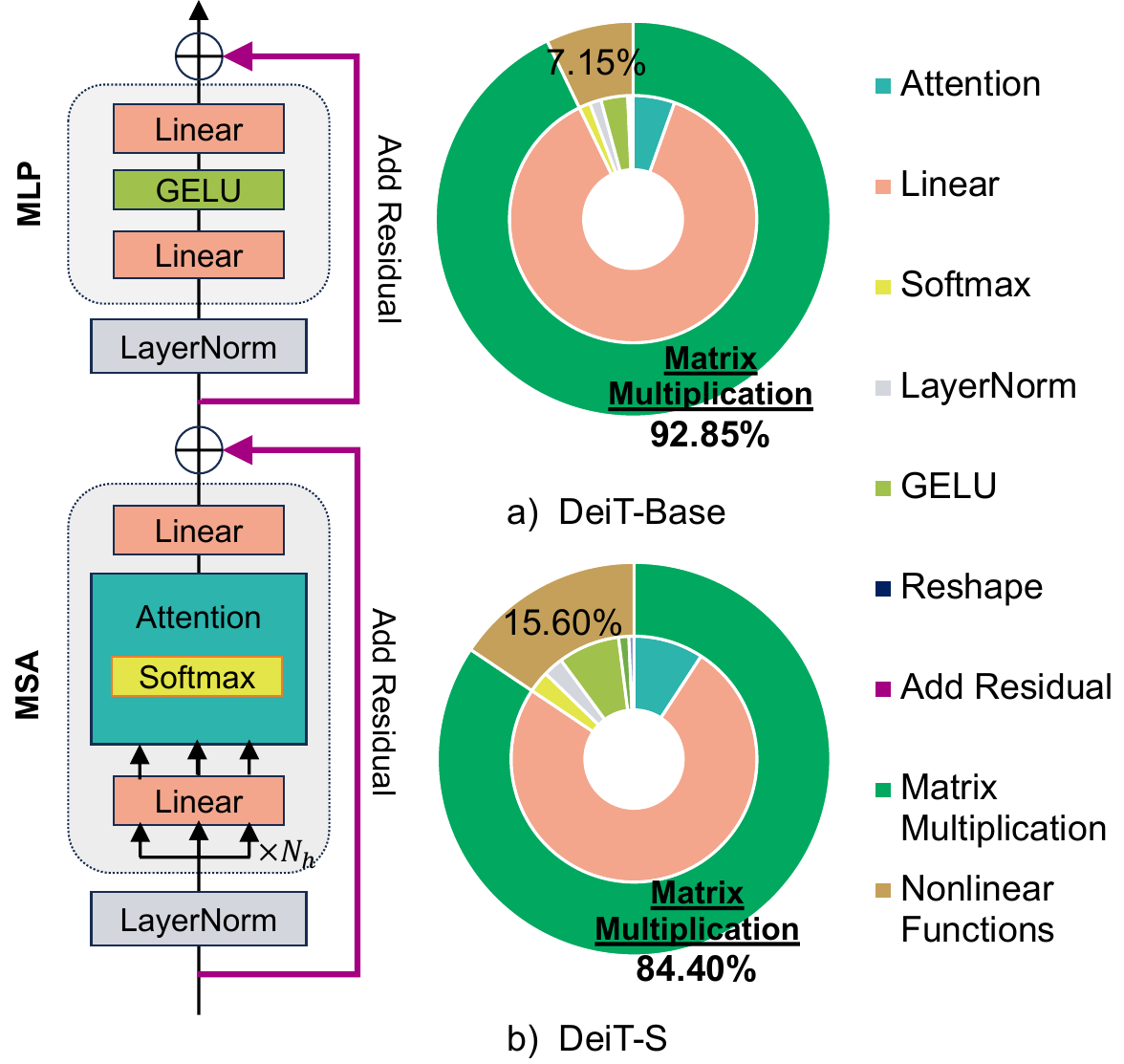}
  \caption{Transformer encoder block structure and ViT model execution performance profiling on CPU for a) DeiT-Base and b) DeiT-S with 12 encoders on ImageNet dataset.
  }
  \label{fig:vit-ori}
\end{figure}

The basic transformer encoder with multi-headed self-attention (MSA) and multi-layer perceptron (MLP) blocks is shown in Figure~\ref{fig:vit-ori}, consisting of multiple different computation components, including linear layer, attention, residual addition, matrix reshape operation, GELU, and layer norm. 
To further understand the bottleneck of the current ViT model structure, we profile the runtime of each component of ViT on a Xeon(R) Silver 4214 CPU~\cite{intel_cpu} using Pytorch Profiler~\cite{pytorch_profiler} as shown in Figure~\ref{fig:vit-ori}. We use the same color to indicate the same component in both the transformer block structure and profiling figures.
It shows matrix multiplication operations dominate the processing time (94.7\% and 87.3\% for DeiT-B~\cite{Touvron2021TrainingDI} and DeiT-S~\cite{deit_s}, respectively) of execution cycles.

Contemporary acceleration methods mainly focus on reducing the practical inference latency of matrix multiplication operations.
They primarily fall into two categories: 1) neural architecture search (NAS) that searches the lighter-weight model; and 2) model compression, especially model quantization that reduces the per-parameter bit-width. However, there are two major challenges
when applying these methods on hardware. 
The first challenge is associated with model quantization. 
It has been revealed that the most suitable quantization schemes/bit-widths depend on model sizes and architectures~\cite{wang2019haq,wu2018mixed}, and there is a vast design space in the quantization of both weights and activations for each layer on different models and hardware. As ViT models become deeper, the design space increases exponentially, resulting in poor performance of rule-based strategies.
Although recent studies explored automated quantization techniques for a given ViT architecture~\cite{wang2019haq,wu2018mixed,uhlich2019mixed}, they did not integrate model quantization with NAS together, which could result in suboptimal performance.
In this paper, we propose the framework of model quantization and NAS co-design for ViTs towards improved performance compared to treating NAS and quantization separately.

The second challenge is the gap between the theoretical computation throughput and the practical inference speed on actual hardware. 
For example, layer-wise (inter-layer) mixed-precision quantization (for CNNs)~\cite{wang2019haq,wu2018mixed} quantizes each layer with a different bit-width and therefore executes layers through distinct hardware components sequentially, leading to low resource utilization.
Furthermore, kernel-wise mixed-precision quantization (for CNNs)~\cite{lou2019autoq} assigns different bit-widths down to the kernel level, significantly diversifying the computing pattern and is inefficient for hardware implementation. 

Recent work FILM-QNN~\cite{sun2022film} and Auto-ViT-Acc~\cite{liauto} leverage the intra-layer mixed quantization to achieve good performance for both model accuracy and throughput on FPGA. By applying two different quantization bit-widths/schemes for different channels and limiting the same mixed-precision ratio across each layer, FPGA can efficiently handle different computations on different hardware resources sharing the same hardware design. However, existing approaches suffer from a manually configured uniform mixed-precision ratio across all layers, potentially compromising quantized model accuracy. Moreover, architectural design considerations are often neglected, limiting the overall model performance.

To address these problems comprehensively, we propose~\M, an integration of a hardware-oriented quantization-aware architecture search targeting ViT. 
First, to fully unleash the computation potential of FPGA resources, we investigate a hardware-friendly row-wise mixed-precision quantization scheme.
At the algorithm level, different from FILM-QNN~\cite{sun2022film} and Auto-ViT-Acc~\cite{liauto}, we quantize different channels within each layer into lower and higher bit-widths with the flexibility of different mix-ratios for layers, which achieves a more fine-grained architecture to maintain the accuracy. 
At the hardware level, we propose the FPGA-based model-adaptive design, including 4-bit atomic computation and hybrid signed/unsigned DSP packing, which set basic hardware units for the lower-bit computation, and decompose the higher-bit computation to lower-bit ones to reuse the resources. % to avoid hardware under-utilization. 
Second, during the supernet training, we propose the mixed-precision weight entanglement mechanism, such that different transformer blocks in subnets can share weights for their common parts in each layer to enable efficient quantization during architecture search and reduce training memory cost. 
On top of that, we establish the corresponding FPGA latency and resource modeling to estimate the inference latency and combine it with an efficient hardware-oriented evolution search method. Based on the above, we integrate with the one-shot NAS algorithm to efficiently find the most accurate quantized model under the given inference latency. 
We also explore the layer scaling in CaiT~\cite{touvron2021going} and extend it to the supernet architecture to improve the training efficiency and model accuracy. To demonstrate the compatibility of our proposed framework with knowledge distillation (KD) and further improve our searched model accuracy, we integrate KD~\cite{hinton2015distilling} into the training process. 
Finally, on the hardware side, we implement the basic computing units for 4-bit weight and 6-bit activations with hybrid signed/unsigned DSP packing optimization to enable efficient FPGA implementation.

The contributions of our work are summarised as follows:
\begin{itemize}
    \item An end-to-end hardware-oriented quantization-aware architecture search framework (\M) for ViTs, achieving superior accuracy and inference speed over prior studies. Latency/resource modeling of the hardware accelerator design is integrated into the search process. 
    \item Hardware-friendly quantization techniques---such as flexible row-wise mixed-precision quantization and mixed-precision weight entanglement---in the architecture search, towards high accuracy, low training cost, and efficient implementation.
    \item Real FPGA implementations of our model-adaptive design, with our proposed 4-bit weight atomic computation and hybrid signed/unsigned DSP packing. 
    \item Integration of proposed supernet layer scaling (SLS) in our framework, achieving further accuracy improvement. Our ablation study also demonstrates our framework's good compatibility with knowledge distillation (KD). 
    % \item  
\end{itemize}

\M~achieves 101.5, 159.6 and 251.6 FPS on the AMD/Xilinx ZCU102 FPGA board with 80.4\%, 78.6\%, and 74.9\% top-1 accuracy for ImageNet, respectively. 

Compared to the representative ViT training-aware quantization~\cite{li2022q} and the post-training quantization~\cite{liu2021post}, at a similar model size, our model achieves 2.1\% and 5.2\% higher top-1 accuracy, respectively. Compared with Auto-ViT-Acc~\cite{liauto}, a state-of-the-art FPGA accelerator for ViT with mixed-scheme quantization (without NAS), we achieve 1.7\% better top-1 accuracy with a similar FPS, and 1.6$\times$ better FPS with a similar level of model accuracy.

\section{Related Work} \label{sec:related}

\subsection{Vision Transformers}

First proposed in \cite{dosovitskiy2020image}, the vision transformer (ViT) is a groundbreaking work that uses transformer blocks for vision tasks. 
Unlike traditional CNN architectures that use a fixed-size window with restricted spatial interactions, 
ViT interprets an image as a sequence of patches and adopts the self-attention mechanism \cite{vaswani2017attention}. This allows all the positions in an image to interact through transformer blocks, which provides the extraordinary capability 
to capture relations at the pixel level in both spatial and temporal domains. However, the original ViT requires pre-training with large-scale datasets such as ImageNet-21k and JFT-300M. To tackle the problem, many variants such as DeiT \cite{Touvron2021TrainingDI} and T2T-ViT \cite{yuan2021tokens} were proposed, which can be well trained with only ImageNet-10k. 
ViTs improve model accuracy at the cost of
increased volume of computation and structural complexity. 
In ViTs, the main model architecture is transformer encoder blocks with multi-headed self-attention (MSA) and multi-layer perceptron (MLP) blocks. These blocks involve large matrix multiplications, which incur the most computational cost. These complex architectures and enormous computation/storage demand make it hard to deploy ViTs on resource-limited edge devices. 

Therefore, we quantize all layers involved in matrix multiplication, but not the non-linear functions, e.g., layer normalization, due to their low computational cost and potential effects on accuracy.

\subsection{Non-Transformer DNN Model Quantization} 
\label{sec:DNN-quant}
\subsubsection{Quantization Scheme}
To compress model size and improve inference speed, model quantization has been widely explored for deep neural networks (DNNs). Existing quantization research can be categorized according to quantization schemes,
such as  binary~\cite{courbariaux2015binaryconnect, rastegari2016xnor}, ternary \cite{he2019simultaneously}, and low-bit-width fixed-point \cite{zhou2016dorefa, choi2018pact,zhou2016dorefa,choi2018pact} quantize models with the same interval between each quantization level.
Although binary and ternary quantization reduce operations and simplify hardware implementation to the extreme, they introduce large accuracy loss due to insufficient bit-width. For example, based on reports from the above works, accuracy typically degrades by $>5\%$ under binary quantization and $2-3\%$ for ternary quantization.
To overcome the large accuracy loss coming from insufficient bit-width, the fixed-point quantization is proposed, applying moderate and adjustable quantization bit-width, to maintain accuracy.
This quantization scheme was implemented with different methods and algorithms, such as DoReFa-Net \cite{zhou2016dorefa} and PACT \cite{choi2018pact}.

Finally, there are also non-linear quantization schemes, such as power-of-two (PoT) \cite{leng2018extremely} and additive PoT \cite{li2020additive}. They replace the multiplication with shifting operations where the distribution of quantization levels becomes unbalanced, having higher precision around the mean and less precision at the two sides.
\subsubsection{Mixed-Precision/Scheme Quantization}
To explore more quantization potential while
preserving the model accuracy, 
Besides the single scheme quantization, some works~\cite{wu2018mixed,dong2019hawq,uhlich2019mixed,wang2019haq,shen2020q} explore inter-layer mixed-precision quantization 
by assigning different precisions to layers. 
For example, HAQ~\cite{wang2019haq} determines the bit-width of each layer by an agent trained with reinforcement learning.
DNAS~\cite{wu2018mixed} used NAS to search layer-wise bit-width.
Furthermore,~\cite{lou2019autoq} explored intra-layer mixed quantization to enable different precisions or schemes within each layer. 
Based on them, hardware designs~\cite{chang2021mix,sun2022film} leveraged the intra-layer mixed-precision/mixed-scheme to enable uniformity within each layer, guaranteeing inference acceleration. However, they need to set the same mixed ratio for layers, which limits the model's accuracy.

\subsection{Transformer and ViT Quantization}
Quantization has also been studied for transformers, especially for natural language processing (NLP) tasks~\cite{zafrir2019q8bert,zhang2020ternarybert,bai2020binarybert}. 
Q8BERT~\cite{zafrir2019q8bert} finetuned BERT through 8-bit quantization-aware training. TernaryBERT~\cite{zhang2020ternarybert} implemented an approximation-based and loss-aware ternary quantization on BERT.
BinaryBERT~\cite{bai2020binarybert} proposed a ternary weight splitting strategy to derive binary BERT with performance as the ternary one. 
Inspired by those,~\cite{liu2021post} and~\cite{liauto} studied quantization on ViT in computer vision tasks. 
PTQ~\cite{liu2021post} evaluated the post-training quantization on ViT and achieved comparable accuracy to the full-precision version. Auto-ViT-acc~\cite{liauto} proposed an FPGA-aware framework with mixed-scheme quantization for ViT, which we will compare in the evaluation. FQ-ViT~\cite{lin2022fq} proposed power-of-two factor and log-int-softmax to proceed with the ViT quantization. Q-ViT~\cite{li2022q} used the switchable scale to achieve head-wise ViT mixed quantization.
However, these works are all based on full-precision pre-trained models and do not include the dimension of network architecture search.

\subsection{Neural Architecture Search}
\subsubsection{NAS Strategies}
There has been a trend to design efficient DNNs with NAS. In general, NAS can be classified into the following categories according to its search strategy.
First, reinforcement learning (RL) methods \cite{zoph2016neural, zhong2018practical, zoph2018learning, baker2016designing,  liu2018progressive, cai2017efficient, pham2018efficient} use recurrent neural networks as predictors validating the accuracy of child networks over a proxy dataset. 
Second, evolution methods \cite{real2019regularized, miikkulainen2019evolving} develop a pipeline of parent initialization, population updating, generation, and elimination of offspring to find desired networks. 
Third, one-shot NAS \cite{bender2018understanding, you2020greedynas, guo2020single} trains a large one-shot model containing all operations and shares the weight parameters with all candidate models. Based on the above work, weight-sharing NAS has become popular due to training efficiency~\cite{yu2020bignas,wang2021attentivenas,sahni2021compofa}. One over-parameterized supernet is trained with weights shared across all sub-networks in the search space. This significantly reduces the computational cost during the search. 
Although most of the above work focuses on the traditional CNN architectures, such as~\cite{yu2020bignas} and~\cite{sahni2021compofa}, 
some works have started investigating the search for efficient ViT networks~\cite{wang2020hat,li2021bossnas,wu2021cvt,chen2021autoformer}. Among them, Autoformer~\cite{chen2021autoformer} entangles the model weights of different ViT blocks in the same layer during supernet training with an efficient weight-sharing strategy to reduce training model storage consumption as well as training time.

\subsubsection{Hardware-Oriented NAS}
Some recent works realize the gap between theoretical computation improvement and practical inference speedup.
They investigate the algorithm/hardware co-design and incorporate the inference latency into NAS \cite{tan2019mnasnet, wu2019fbnet, li2021npas}, which is more accurate than intuitive volume estimation by MAC operations. For example, MnasNet \cite{tan2019mnasnet} and NPAS~\cite{li2021npas} utilize the latency on mobile devices as the reward to perform RL search, where gradient-based NAS work FBNet~\cite{wu2019fbnet} adds a latency term to the loss function. 
However, these works neither target ViTs nor exploit quantization in the hardware-aware ViT search.

\section{Hardware-Oriented Quantization-Aware NAS for
ViTs} \label{sec:design}

\subsection{Row-Wise Flexible Mixed-Precision Quantization with Hardware/Software Co-design}

\begin{figure}[!tb]
  \centering
  \includegraphics[width=0.95\linewidth]{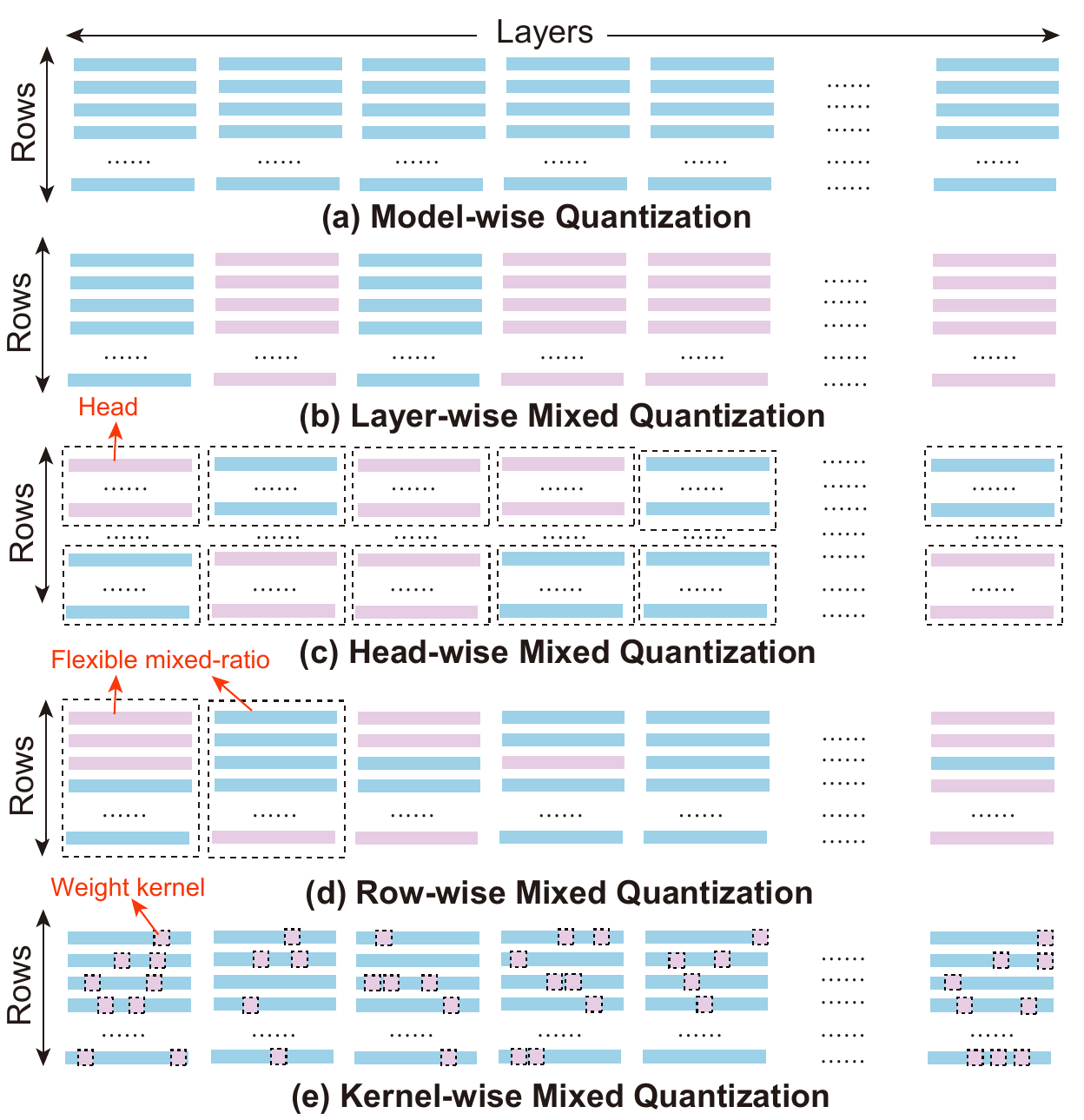}
  \caption{Comparison of mixed-precision quantization under different granularity. We use the example of two different bit-widths, represented as blue and pink colors. We propose the row-wise flexible mixed-precision quantization in~(d).}
  \label{fig:quant}
\end{figure}

\begin{figure*}[htb]
  \centering \includegraphics[width=1.0\linewidth]{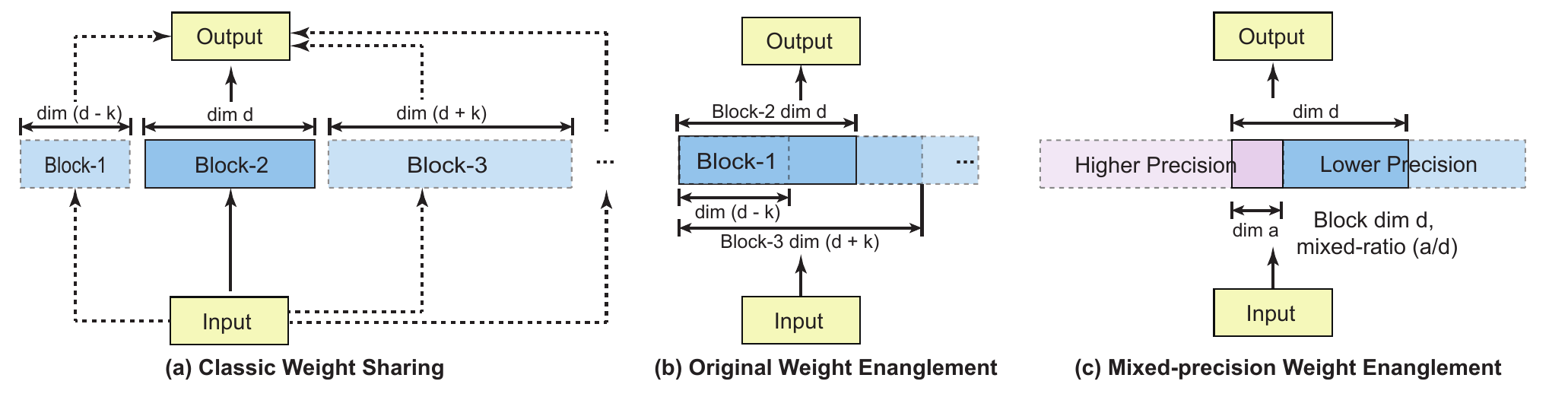}
  \caption{Classic weight sharing and original weight entanglement versus our proposed mixed-precision weight entanglement.}
  \label{fig:entanglement}
\end{figure*}

Figure~\ref{fig:quant} classifies quantization with different levels of granularity. Model-wise quantization~\cite{zhou2016dorefa, choi2018pact} uses a unified quantization bit-width for the whole model and thus misses some quantization opportunities.
On the other hand, mixed-precision quantization, as discussed in related work, explores more quantization potential (i.e., quantizing each component to a bit-width as low as possible) while preserving the accuracy of the model.
Specifically, layer-wise (inter-layer) mixed-precision quantization~\cite{wang2019haq,wu2018mixed} sets each layer with a specific quantization bit-width. 
Besides that, Q-ViT~\cite{li2022q} proposed a head-wise mixed-precision quantization scheme, which assigns different bit-widths to different attention heads. Both the layer-wise and head-wise quantization schemes suffer from limited quantization flexibility without considering the variance inside each layer. 
Moreover, fixed row-wise (intra-layer) mixed-precision quantization is proposed in prior work~\cite{sun2022film}, which uses different quantization bit-widths for different channels in each CNN layer and limits the same mixed-precision ratio across different CNN layers, and thus multiple layers can share the same hardware design, making it more hardware-friendly. Finally, kernel-wise mixed-precision quantization~\cite{lou2019autoq} assigns different quantization bit-widths down to the kernel level, which greatly diversifies the computing pattern and makes it inefficient to implement on hardware.

Based on the above discussion, we use the row-wise flexible mixed-precision quantization scheme for ViTs, as shown in Figure~\ref{fig:quant}(d), which preserves the quantization flexibility among layers for better accuracy while maintaining the hardware uniformity for more efficient implementation. Different from~\cite {sun2022film} that limits the same mixed-precision ratio across CNN layers, for ViTs, we have to provide the flexibility to obtain different mixed ratios in different layers to maintain the model accuracy. 
To maintain hardware uniformity and avoid hardware under-utilization, we propose to design the basic hardware units for the lower-bit computation, decompose the higher-bit computation into lower-bit ones, and reuse the basic hardware units (described in Section~\ref{subsec:hardware_unification} and Section~\ref{subsec:dsp_packing}).
As a result, we have preserved the uniformity of the hardware design and enabled the flexible bit-width mixed-ratio among ViT layers. We explain the hardware details in Section~\ref{sec:design_fpga}.

\subsection{Intra-layer Mixed-Precision Weight Entanglement}

In classical one-shot NAS, the weights of each sample candidate are shared with the supernet during training. However, as shown in Figure~\ref{fig:entanglement} (a), when using the classical weight-sharing strategy, the building blocks from multiple subnets, even in the same layer, are isolated. Therefore, it leads to higher memory costs and slower training convergence.

To address this problem, weight entanglement is proposed  in~\cite{chen2021autoformer} to reduce the supernet model size: as shown in Figure~\ref{fig:entanglement} (b), different transformer blocks can share their common weights in each layer. 
It also allows each block to be updated more times than the previous independent training strategy, thus achieving faster convergence.
However, this structure is hard to combine with mixed quantization since one shared weight cannot be trained into two different bit-widths at the same time (i.e., bit-width conflict).

In this paper, we propose the mixed-precision weight entanglement, as shown in Figure~\ref{fig:entanglement} (c),
to incorporate the quantization search while preventing the potential bit-width conflicts problem in the shared weight.
Mixed-precision weight entanglement block contains two parts of weights with different precisions. Different quantization mixed ratios can be achieved by extracting different percentages of weights over these two parts.
In the implementation, for each layer, we only need to store the weights of the largest of the $n$ homogeneous candidate blocks with both the 4-bit and 8-bit weights. The remaining smaller building blocks can extract the weights directly from the largest building block. The different quantization mixed ratios can be reached by moving the position of the selected block.

\subsection{Supernet Layer Scaling (SLS) Structure}

The layer scaling structure proposed by CaiT~\cite{touvron2021going} improves the stability of the optimizations when training transformers for image classification, thus improving the model accuracy. We explore this structure and extend it to supernet layer scaling (SLS). 

Layer scaling is a per-channel multiplication of the vector produced by each residual block. For ViT, this layer is deployed after the multi-head self-attention (MSA) and multi-layer perceptron (MLP) modules in each encoder block. The objective is to group the updates of the weights associated with the same output channel. 
Layer scaling can be denoted as a multiplication by a diagonal matrix $\operatorname{diag}\left(\lambda_{l, 1}, \ldots, \lambda_{l, d}\right)$
on the output of $l$-th residual block, where $d$ is the corresponding number of output channels in the model. All $\lambda$s are learnable weights. 

\begin{figure}[h]
  \centering \includegraphics[width=0.75\linewidth]{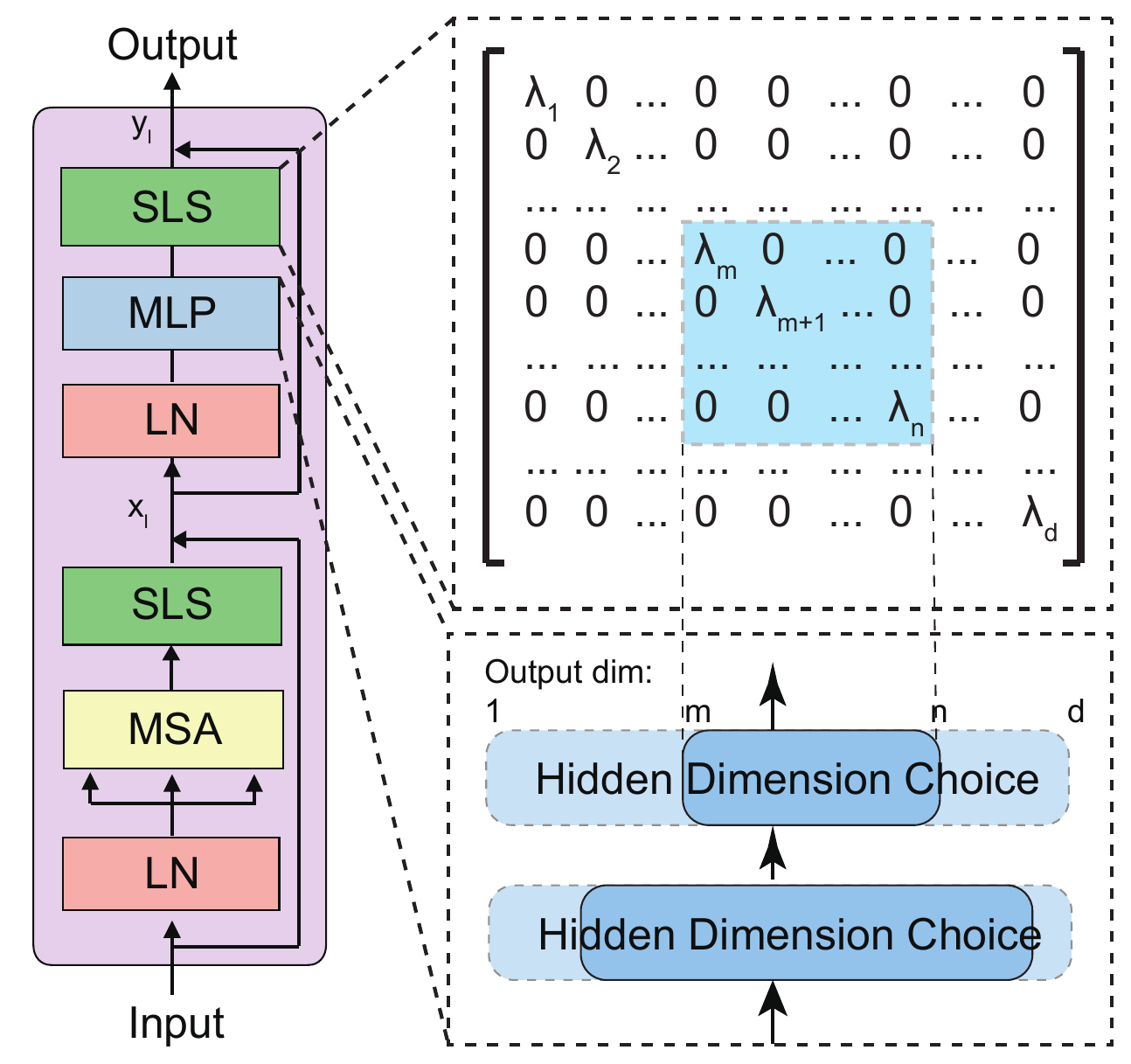}
  \caption{Supernet layer scaling (SLS) in \M~encoder block. We use the SLS after MLP as an example.}
  \label{fig:layerscale}
\end{figure}

To fit our mixed-precision weight entanglement strategy, different from the original CaiT~\cite{touvron2021going} implementation that uses the whole layer scaling in every training iteration, our SLS extracts the corresponding elements synchronized with the output dimension of the selected subnet while keeping the other weights frozen. As shown in Figure~\ref{fig:layerscale}, using the residual block of MLP as an example, assuming that the current MLP's output dimension starts from $m$-th channel and ends at $n$-th channel, the supernet layer scaling computation can be formulated as:
\begin{equation}
\begin{aligned}
y_{l} &=x_{l}+\operatorname{diag}\left(\lambda_{l, m}, \ldots, \lambda_{l, n}\right) \times \operatorname{MLP}\left(\operatorname{LN}\left(x_{l}\right)\right),
\end{aligned}
\end{equation}
where $x_{l}$ and $y_{l}$ denote the input and output, respectively; LN means the layer normalization.

\begin{figure*}[!tb]
  \centering
  \includegraphics[width=1.0\linewidth]{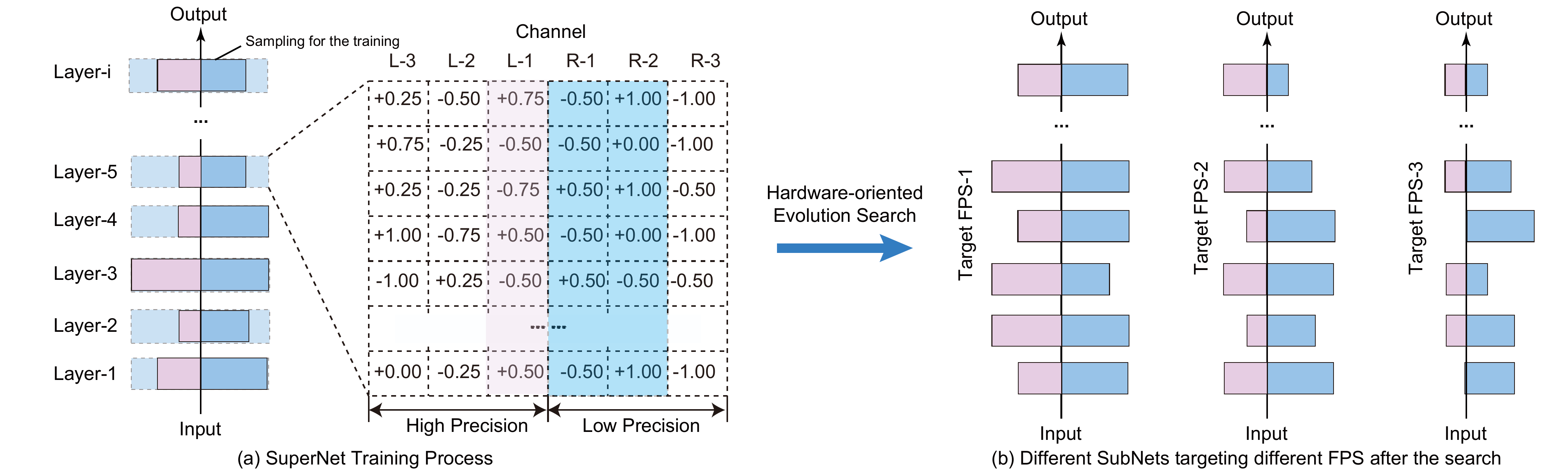}
  \caption{SuperNet training process including candidate sampling and the searched results for different targeting FPS. We use a model with layers of 6 channels as a toy example.}
  \label{fig:toy}
\end{figure*}

\subsection{End-to-End \M~Framework}

\subsubsection{One-shot NAS Algorithm}
Our one-shot NAS algorithm consists of two steps:
\begin{itemize}
\item We train a supernet to directly sample different quantized architectures as child models for training. The supernet is trained with SLS and KD techniques. The search space is encoded in the supernet, and the parameters of all candidate networks in the search space are optimized simultaneously by our proposed weight-sharing during training.
\item We select architectures from the pre-trained supernet using the hardware-oriented evolution search method to find the most accurate model under the given hardware resource constraints. We search based on the hardware latency/FPS and resource modeling illustrated in Section~\ref{subsec:hardware_models}.
\end{itemize}

Here, we show a toy example of the supernet training process including candidate sampling and the corresponding searched results for different targeting FPS in Figure~\ref{fig:toy}. Figure~\ref{fig:toy} (a) illustrates one iteration of the supernet training process, where the pink area indicates the sampled high precision values and the blue area indicates the sampled low precision values in the supernet. The light blue area indicates the frozen values (currently not sampled) in this iteration. After the supernet training and the hardware-oriented evolution search, we could obtain different models targeting different frames per second (FPS) as shown in Figure~\ref{fig:toy} (b). For the sake of brevity, we only show the quantized value here. The scaling factor along with other related structures is omitted.

\subsubsection{Search Space}
We show our search space design in Table~\ref{tab:search_space}. Our search components include the overall embedding dimension, the number of transformer layers, the quantization mixed-ratio (i.e., the percentage of 8-bit weights mixed in the layer) for each linear layer, and the hidden dimension and expansion ratio (for MLP) in each ViT encoder block.

To accelerate the supernet training process and improve the overall model performance, we partition the large-scale search space into two sub-spaces and encode them into two independent supernets for QUASAR-Small and QUASAR-Large, respectively. By splitting and customizing search space for supernets of different sizes, we mitigate the training interference caused by the huge subnets’ difference. This training strategy has been proved in~\cite{zhao2021few}. Such partition allows the search algorithm to concentrate on finding models within a specific hardware inference latency, which can be specialized by users according to their available resources and application requirements.  It also reduces gradient conflicts between large and small sub-networks trained via weight-sharing due to gaps in model sizes.

\begin{table}[h]
\caption{An illustration of our search space: It is divided into two independent supernets within the different parameter ranges to satisfy different resource constraints.}\label{tab:search_space}
\centering
\resizebox{0.95 \columnwidth}{!}{
\begin{tabular}{c|ccc }
\toprule
    & QUASAR-Small & QUASAR-Large   \\
\hline
Embed Dimension&  (192, 216, 240)  &  (320, 384, 448)   \\
Hidden Dimension& (192, 256)  &  (320, 384, 448)   \\
8-bit Mixed-ratio & (0\%, 25\%, 50\%) & (0\%, 25\%, 50\%)\\
Expansion Ratio& (3.5, 4)  & (3, 3.5, 4)   \\
Number of Layers&  (12,13,14)  & (12,13,14)\\

\bottomrule
\end{tabular}
}
\end{table}

\subsubsection{Supernet Training}
In each iteration, we randomly select a quantized ViT architecture from the search space. Then we obtain its weights from the supernet and compute the losses of the subnet. Finally, we update the corresponding weights with the remaining weights frozen.  The architecture search space $P$ is encoded in a supernet denoted as $\mathcal{S} (P, W_P)$, where $W_P$ is the weight of the supernet that is shared across all the candidate architectures. Algorithm~\ref{alg:supernet_train} illustrates the training procedure of our supernet.

\begin{algorithm}[tb]
\begin{algorithmic}
   \caption{Supernet Training.} 
   \label{alg:supernet_train}
   \STATE {\bfseries Input:} Training epochs $N$, search space $\mathcal{P}$, supernet $\mathcal{S}$, loss function $L$, train dataset $D_{train}$, initial supernet weights $\mathcal{W}_{\mathcal{P}}$, candidate weights $\mathcal{W}_p$
   \FOR{$i$ in $N$ epochs} 
   \FOR{data, labels in $D_{train}$ } 
   \STATE{Randomly sample one quantized ViT architecture from search space $\mathcal{P}$}
   \STATE{Obtain the corresponding weights $\mathcal{W}_p$ from supernet $\mathcal{W}_{\mathcal{P}}$}
   \STATE{Compute the gradients based on $L$}
  \STATE{ Update the corresponding part of $\mathcal{W}_p$ in $\mathcal{W}_{\mathcal{P}}$ while freezing the rest of the supernet $\mathcal{S}$}
   \ENDFOR
   \ENDFOR 
   \STATE{\bfseries Output $\mathcal{S}$}
\end{algorithmic}
\end{algorithm}

\subsection{Hardware-Oriented Evolution Search}
In our hardware-oriented evolution search for crossover, two random candidate architectures are first picked from the top candidates. Then we uniformly choose one block from them in each layer to generate a new architecture. For mutation, a candidate mutates its depth with probability $P_d$ first. Then it mutates each block with a probability of $P_m$ to produce a new architecture. Newly produced architectures that do not satisfy the constraints will not be added for the next generation. To evaluate the candidates, we perform hardware latency and resource modeling based on the proposed row-wise flexible mixed-precision quantization scheme. The details of the modeling have been discussed in Section~\ref{subsec:hardware_models}.

\subsection{Integration with Knowledge Distillation (KD)}

To demonstrate the compatibility of our proposed framework with knowledge distillation (KD) and further improve the accuracy of our supernet,
we also integrate KD~\cite{hinton2015distilling} in our training process. We use the pre-trained RegNetY-32G \cite{radosavovic2020designing} with 83.6\% top-1 accuracy as different teacher models. We also apply the soft distillation method.  
Soft distillation \cite{hinton2015distilling} minimizes the Kullback-Leibler divergence between the softmax of the teacher and the softmax of the student model. The distillation loss is:
\begin{equation}
L_{soft} = (1- \alpha)L_{CE}(\psi (Z_s),y) + \alpha \tau ^2 L_{KL} (\psi (\frac{Z_s}{\tau}), \psi (\frac{Z_t}{\tau})),
\label{eq:soft_dis}
\end{equation}
where $Z_t$ and $Z_s$ are the logits of the teacher and student models, respectively. $\psi$ is the softmax function. $\tau$ is the temperature for the distillation, $\alpha$ is the coefficient balancing the Kullback–Leibler divergence loss ($L_{KL}$), and the cross-entropy ($L_{CE}$) on the ground truth labels $y$ in the distillation.

\section{FPGA Hardware Design for \M} \label{sec:design_fpga}

\subsection{Overall Hardware Design for~\M{}} \label{subsec:hardware_design}

\begin{figure}[!htb]
\centerline{\includegraphics[width=0.95\columnwidth]{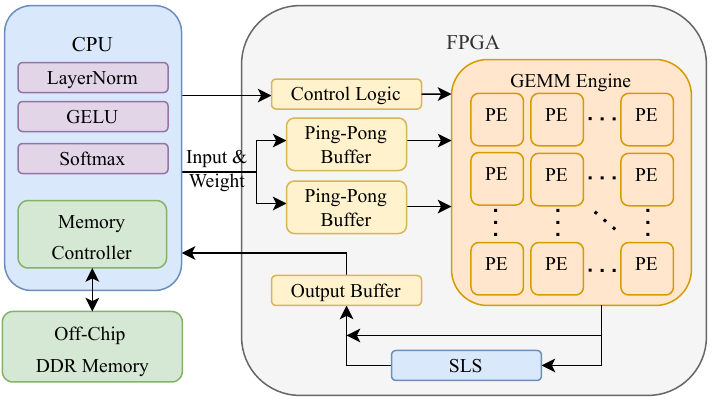}}
    \caption{\M{} hardware architecture.}
    \label{fig:fpga_arch}
% \vspace{-0.4cm}
\end{figure}

Figure~\ref{fig:fpga_arch} presents the overall hardware architecture of the~\M{} accelerator on the ARM-FPGA platform. Below is how each module in 
ViT is mapped to the hardware in Figure~\ref{fig:fpga_arch}. 
The most time-consuming MSA and MLP modules are accelerated by our GEMM engine on the FPGA, which is similar to the recent Auto-ViT-Acc work~\cite{liauto}. 
The lightweight SLS modules right after MSA and MLP layers are also accelerated on the FPGA to avoid time-consuming execution on the ARM CPU. 
The less time-consuming modules including layer normalization and activation functions (i.e., Softmax or GELU) are executed on the ARM CPU, due to their complex structure for FPGA implementation. 
The hardware engines on the FPGA and software modules on the ARM CPU exchange data via the shared off-chip memory.

As previously mentioned, we mainly focus on the most time-consuming GEMM engine design. Due to the limited on-chip memory capacity and computing resource on the FPGA, for each ViT layer (i.e., MSA and MLP), our GEMM engine processes the input, weight, and output data in tiles: a small tile of the input (tokens) and weight of each ViT layer are first loaded from the off-chip DDR memory to the on-chip buffers, then they are processed by the GEMM engine all on-chip. To improve the performance, the double buffering technique is applied again to overlap the off-chip memory accesses and GEMM computation, shown in Figure~\ref{fig:fpga_arch}. 

Next, we present our design of the basic hardware units in the GEMM engine and the corresponding DSP (digital signal processor) packing optimization, as well as the hardware resource and latency modeling for the tiled GEMM design.

\subsection{Unification of Atomic Computation} \label{subsec:hardware_unification}
One major challenge in the FPGA accelerator design is to efficiently support flexible mixed ratios of different bit-width computations across ViT layers.
On one hand, putting multiple copies of hardware accelerator designs for each mixed-ratio (i.e., each layer) simultaneously on the FPGA leads to severe hardware resource contention and under-utilization, since layers are executed sequentially. 
On the other hand, pre-synthesizing multiple copies of hardware accelerator designs for each layer and reconfiguring the FPGA for each layer incurs significant FPGA reconfiguration overhead.

Inspired by the approach proposed in QGTC~\cite{wang2022QGTC} to support arbitrary bit-width computation for quantized graph neural networks on GPUs, in our FPGA hardware design, we unify the basic processing elements to process 4-bit weight atomic computations and construct the 8-bit weight data computations using two 4-bit weight data operations as such: for multiplication between an N-bit activation value ($act_N$) and an 8-bit weight value ($wgt_8$), 
we derive the corresponding product as:
\begin{equation}
\begin{aligned}
&  act_{N} \cdot wgt_{8} = act_{N} \cdot wgt_{h4} << 4 + act_{N} \cdot wgt_{l4},
\end{aligned}
\end{equation}
where $wgt_{h4}$ and $wgt_{l4}$ represent the higher and lower 4-bit data of $wgt_8$, respectively.
The multiplication result between $act_{N}$ and $wgt_{h4}$ are left shifted by 4 bits.

Based on this unification, we propose hybrid signed/unsigned DSP packing to handle the 4-bit weight atomic computation.

\subsection{Proposed Hybrid Signed/Unsigned DSP Packing} \label{subsec:dsp_packing}
To fully exploit the potential of DSP resources on FPGAs, we pack multiple low-bit multiplications within each DSP block following~\cite{Xilinx-dspPack-Int8,Xilinx-dspPack-Int4}.
Each DSP block (DSP48E2) on the AMD/Xilinx ZCU102 FPGA board could support the computation of $P{=}(A{+}D) {\times} B$, where both $A$ and $D$ are 27-bit operands, $B$ is an 18-bit operand, and $P$ is the 45-bit output.
In our study, we explore the following two DSP packing schemes and discuss their design trade-offs. The activation bit-width $N$ is set to 6 to fully exploit the DSP for computation.

\begin{figure*}[h]
  \centering
  \includegraphics[width=1.0\linewidth]{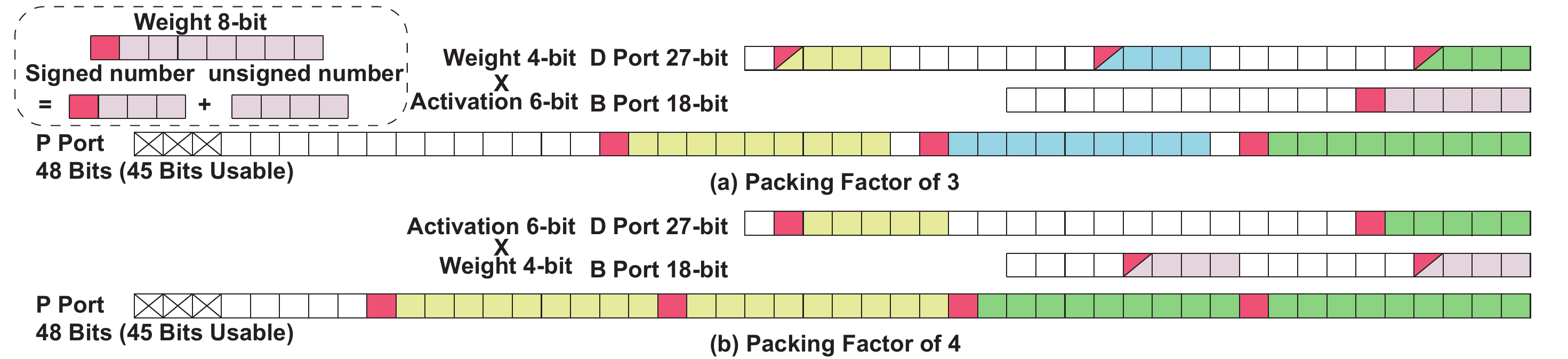}
  \caption{Illustration of DSP multiplication packing schemes for (a) 3 weights sharing 1 activation with packing factor of 3, and (b) 2 weights sharing 2 activations with packing factor of 4.}
  \label{fig:dsp-pack}
  % \vspace{-0.4cm}
\end{figure*}

\begin{itemize}
    \item \textbf{Packing factor 3 (3 weights sharing 1 activation).}
    In Figure~\ref{fig:dsp-pack} (a), three $4 \times 6$-bit multiplications are packed into a single DSP block, by holding one 6-bit signed activation in port $B$ and three 4-bit weight values in port $D$. To pack three weights into a single 27-bit port $D$, look-up tables (LUTs) are utilized to first combine two weights and then integrate them with the third weight data. 
    With this DSP packing scheme, for the W4A6 (i.e., 4-bit weight and 6-bit activation) computation, we could pack three 4-bit weights that share the same activation. And for the W8A6 computation, we could use two DSPs to process the upper and lower 4-bit of three 8-bit weights in parallel. Note that after the 8-bit weights are decomposed into two 4-bit weights, only the upper 4-bit weights contain the sign bits and should be sign-extended (required by DSP for output correctness~\cite{Xilinx-dspPack-Int8,Xilinx-dspPack-Int4}), the lower 4-bit weights should be treated as unsigned values.
    \item  \textbf{Packing factor 4 (2 weights sharing 2 activations).}
    In Figure~\ref{fig:dsp-pack} (b), four $4 \times 6$-bit multiplications are packed into a single DSP block, by holding two 6-bit signed activation values in port $D$ and two 4-bit weights in port $B$. It is worth mentioning the port placement of the activation and weight values are swapped from the packing factor 3 scheme, to increase the packing strength.
    With this DSP packing scheme, for the W4A6 computation, we could pack a pair of two 4-bit weights that share the same pair of activation values. And for the W8A6 computation, a similar technique as the previous packing scheme is used to separately handle the upper and lower 4-bit values of an 8-bit weight. Again for the two decomposed 4-bit weights, only the upper 4-bit weights contain the sign bit and should be sign-extended (required by DSP for output correctness~\cite{Xilinx-dspPack-Int8,Xilinx-dspPack-Int4}), but the lower 4-bit weights should be treated as unsigned values.
\end{itemize}

% \begin{comment}
\subsection{Hardware Resource and Latency Modeling} \label{subsec:hardware_models}

Here, we present our hardware resource and latency modeling used in the hardware-oriented evolution search.

\begin{table}[htb]
\tabcolsep 3pt
\centering
\small

\caption{Notations for~\M{} accelerator}
\label{tab:notation}

\scalebox{1.0}
{
\begin{tabular}{p{1.5cm}|p{6.0cm}}
\toprule
Notation & Description \\
\midrule

$M$ ($N$) & Number of output (input) channels \\ \hline
$F$ & Number of token sequences \\ \hline
$T_n$ & Tiling size for data in input channel dimension for each head \\ \hline
$T_m$ & Tiling size for data in output channel dimension \\ \hline
$N_h$ & Total number of heads \\ \hline
$P_F$ & Parallel factor along the number of tokens \\ \hline
$D_{act}$  & Number of data packed as one for activations  \\ \hline
$D_{wgt}$ & Number of data packed as one for weights \\ \hline
$A_{\mathrm{in}}$ ($A_{\mathrm{out}}$, $A_{\mathrm{wgt}}$) & Number of AXI ports used for data transfer of input (output, weight) tile \\
 \hline
$L_{\mathrm{in}}$ ($L_{\mathrm{wgt}}$, $L_{\mathrm{out}}$, $L_{\mathrm{cmpt}}$)  & Number of clock cycles for input transfer (weight transfer, output transfer, computation) for a tile \\
\hline
$S_{\mathrm{dsp}} (S_{\mathrm{lut}})$ & Available number of DSPs (LUTs) on FPGA \\ \hline
$C_{\mathrm{dsp}}$ & DSP cost for each MAC operation \\  \hline
$C_{\mathrm{lut}}$ & LUT cost for each MAC operation \\ \hline
$C_{lut}^{dsp}$ & Number of LUTs used by a multiplication executed on DSPs \\ \hline
$N_{dsp}$  & Number of multiplication executed on DSPs \\ \hline
$N_{lut}$ & Number of multiplication executed on LUTs \\ \hline
$N_{tot}$  & The total number of multiplication on FPGA \\ \hline
$\gamma_{dsp}$ ($\gamma_{lut}$) & DSP (LUT) utilization threshold  \\
\hline
$f$ & FPGA accelerator frequency  \\
\hline
$FPS$ & Frames per second  \\
\bottomrule
\end{tabular}
}
\end{table}

\subsubsection{Resource Modeling}
To help guide the neural architecture search, we provide details of the resource and latency models of our FPGA accelerator design (mainly the GEMM engine). 
Table~\ref{tab:notation} lists the notations used in our models.
We design our FPGA accelerator to fully leverage the available FPGA computing resources (i.e., DSPs and LUTs), on-chip memory (i.e., BRAMs), and off-chip memory bandwidth. 
To fully exploit the computing capability with the available hardware resources, We maximize the total number of parallel basic hardware compute units using both DSPs (i.e., $N_{dsp}$) and LUTs (i.e., $N_{lut}$) for the datapath of our accelerator as
\begin{equation}\label{eqn:max}
\begin{aligned}
& N_{tot} = \mathop{\mathrm{maximize}} \left\{ N_{dsp} + N_{lut}\right\},
\end{aligned}
\end{equation}
while satisfying the following resource constraints
\begin{equation}\label{eqn:C1}
\begin{aligned}
&  N_{dsp} \cdot C_{dsp}  \leq S_{dsp} \cdot \gamma_{dsp},
\end{aligned}
\end{equation}
\begin{equation}\label{eqn:C2}
\begin{aligned}
&  N_{lut} \cdot C_{lut} + N_{dsp} \cdot C_{lut}^{dsp} \leq S_{lut} \cdot \gamma_{lut},
\end{aligned}
\end{equation}
where constraints~\ref{eqn:C1} and~\ref{eqn:C2} bound the DSP and LUT utilization to be under the allowable thresholds, i.e., $\gamma_{dsp}$ and $\gamma_{lut}$ with the total resource amounts denoted by $S_{dsp}$ and $S_{lut}$. The hardware costs of each multiplication implementation are denoted as $C_{lut}$ and $C_{lut}^{dsp}$. 

In order to choose the best implementation method for the basic hardware units of our design, we characterize the FPGA resource consumption using Xilinx Vivado 2020.1~\cite{url-vivado} for the three cases in Table~\ref{tbl:cu_resource}, i.e., (a) multiplications executed on DSPs with packing factor of 3, (b) multiplications executed on DSPs with packing factor of 4, and (c) multiplications executed purely using LUTs. 
As observed in Table~\ref{tbl:cu_resource}, we derive the hardware costs of each multiplication implementation, particularly, the LUT costs for pure-LUT-based and DSP-based methods correspond to $C_{lut}$ and $C_{lut}^{dsp}$. 
Note that the DSP-based approach also consumes LUTs, due to data packing on the input operands and output data construction operations, such as bit shifting and data accumulation. 
\rev{
Regarding the efficiency lost by decomposing 8-bit to 4-bit, for the composed W8A6 computation, on average, we can achieve one computation with 25.8 LUTs and 0.5 DSP or purely 66.7 LUTs. In contrast, direct W8A6 computation used in~\cite{sun2022film} (i.e., packing two W8A6 operations within one DSP method) requires 21.5 LUTs and 0.5 DSP or purely 62.2 LUTs. Since most of the weights are in 4-bit, using decomposing does not affect the overall performance much by a slight increase in the LUT utilization. 
In terms of the efficiency of DSP packing, a single DSP can pack four W4A6 operations at most theoretically, which is achieved by our approach.
}

For the LUT usage, shown in Table~\ref{tbl:cu_resource}, we have:
\begin{equation}\label{eqn:C3}
\begin{aligned}
&   C_{lut}^{dsp,pack3} \textless C_{lut}^{dsp,pack4} \textless C_{lut}.
\end{aligned}
\end{equation}

\begin{table}[!tb]
\caption{\rev{FPGA resource consumption for a single DSP-based or LUT-based basic hardware compute unit. W4A6 denotes 4-bit weight and 6-bit activation; W8A6 denotes 8-bit weight and 6-bit activation.}}
\label{tbl:cu_resource}
\centering
\begin{tabular}{c|c|c|cc}
\hline
\multirow{2}{*}{} &
  \multirow{2}{*}{} &
  \multirow{2}{*}{pure LUT-based} &
  \multicolumn{2}{c}{DSP-based} \\ \cline{4-5} 
 &
   &
   &
  \multicolumn{1}{c|}{\begin{tabular}[c]{@{}c@{}}packing\\ factor 3\end{tabular}} &
  \begin{tabular}[c]{@{}c@{}}packing\\ factor 4\end{tabular} \\ \hline
\multirow{2}{*}{W4A6} & $C_{LUT}$ & 33.3 & \multicolumn{1}{c|}{10.9} & 12.9 \\ \cline{2-5} 
                      & $C_{DSP}$ & 0    & \multicolumn{1}{c|}{0.33} & 0.25 \\ \hline
\multirow{2}{*}{\begin{tabular}[c]{@{}c@{}}W8A6\\ \end{tabular}} &
  $C_{LUT}$ &
  66.7 &
  \multicolumn{1}{c|}{21.9} &
  25.8 \\ \cline{2-5} 
                      & $C_{DSP}$ & 0    & \multicolumn{1}{c|}{0.67} & 0.5  \\ \hline
\multirow{2}{*}{\begin{tabular}[c]{@{}c@{}}W8A6\\ (direct)\end{tabular}} &
  $C_{LUT}$ &
  62.2 &
  \multicolumn{2}{c}{21.5} \\ \cline{2-5} 
                      & $C_{DSP}$ & 0    & \multicolumn{2}{c}{0.5}          \\ \hline
\end{tabular}
\end{table}

In the final implementation, regarding which DSP packing scheme to use and whether to use the pure-LUT-based method for the basic hardware compute units, there are several situations according to the available FPGA resources.

\textbf{Situation-1:} When $S_{lut}$ is limited and insufficient to hold the LUT consumption of full utilization of DSP packing with a factor of 3, denoted as:
\begin{equation}\label{eqn:C6_2}
\begin{aligned}
 S_{lut} \cdot \gamma_{lut} \leq 3 \cdot  S_{dsp} \cdot \gamma_{dsp} \cdot C_{lut}^{dsp,pack3}.
% 3 \cdot  S_{dsp} \cdot \gamma_{dsp} \cdot C_{lut}^{dsp,pack3} \leq S_{lut} \cdot \gamma_{lut},
\end{aligned}
\end{equation}

In this case, to fully utilize the computation resources, we directly allocate DSP-based computations with a packing factor of 3 as much as possible until we reach the LUT resource limit.

\textbf{Situation-2:} When $S_{lut}$ is enough to hold all the LUT consumption from DSP packing with the factor of 4, satisfying:
\begin{equation}\label{eqn:C4}
\begin{aligned}
4 \cdot  S_{dsp} \cdot \gamma_{dsp} \cdot C_{lut}^{dsp,pack4} \leq S_{lut} \cdot \gamma_{lut}.
\end{aligned}
\end{equation}

Based on Eq.~\ref{eqn:max}, \ref{eqn:C1}, \ref{eqn:C2} and \ref{eqn:C4}, we can conclude that using DSP packing with the factor of 4 is more efficient if it meets the following condition:
\begin{equation}\label{eqn:C5}
\begin{aligned}
&  (4 \cdot C_{lut}^{dsp,pack4} - 3 \cdot C_{lut}^{dsp,pack3}) \cdot  S_{dsp} \cdot \gamma_{dsp} \leq S_{lut} \cdot \gamma_{lut} \cdot C_{lut}.
\end{aligned}
\end{equation}

If it does not satisfy Eq.~\ref{eqn:C5},
we perform DSP-based computations with a packing factor of 3. Besides that, for the remaining available LUT resources, pure-LUT-based computation is also conducted in both cases.

\textbf{Situation-3:} When $S_{lut}$ is between the LUT demands by fully deploying DSP computations with packing factors of 3 and 4, satisfying:
\begin{equation}\label{eqn:C6}
\begin{aligned}
 3 \cdot  S_{dsp} \cdot \gamma_{dsp} \cdot C_{lut}^{dsp,pack3} \leq S_{lut} \cdot \gamma_{lut} \leq  \\
 4 \cdot  S_{dsp} \cdot \gamma_{dsp} \cdot C_{lut}^{dsp,pack4}. 
\end{aligned}
\end{equation}

Based on Eq.~\ref{eqn:max}, \ref{eqn:C1}, \ref{eqn:C2}, and \ref{eqn:C6}, we can conclude that using DSP packing with the factor of 4 is more efficient if it meets the following condition:
\begin{equation}\label{eqn:C5_2}
\begin{aligned}
& \frac{S_{lut} \cdot \gamma_{lut} + 3 \cdot S_{dsp} \cdot \gamma_{dsp} \cdot (C_{lut} - C_{lut}^{dsp,pack3})}{C_{lut}} \leq \frac{S_{lut} \cdot \gamma_{lut}}{C_{lut}^{dsp,pack4}},
\end{aligned}
\end{equation}

In this case, we conduct only DSP-based computations with a packing factor of 4. Otherwise, if it does not satisfy Eq.~\ref{eqn:C5_2}, we perform DSP-based computations with a packing factor of 3 and pure LUT-based computation for the remaining available LUT resource.

\subsubsection{Latency Modeling}

\begin{figure}[!tb]
  \centering
  \includegraphics[width=0.8\linewidth]{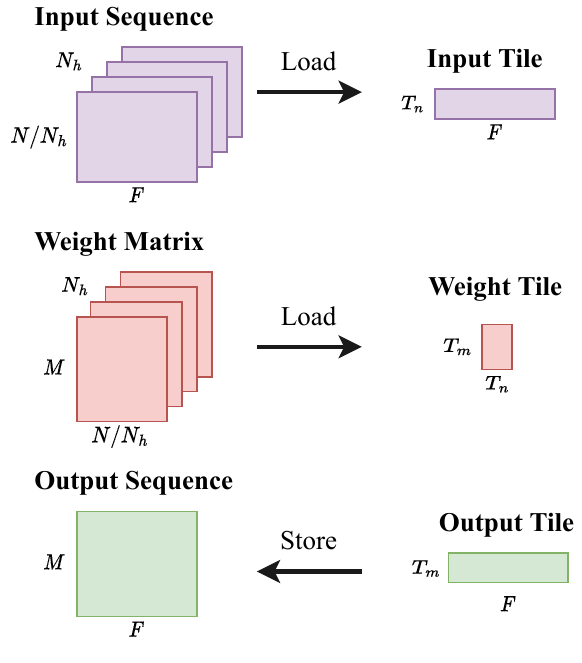}
  \caption{Data tiling in ViT computations.}
  \label{fig:tiling}
\end{figure}

As discussed in 
the Hardware Design Section, 
our GEMM hardware engine accelerates the major MSA and MLP modules and processes their input, weight, and output data in tiles, as shown in Figure~\ref{fig:tiling}. 
our GEMM hardware engine accelerates the major MSA and MLP modules and processes their input, weight, and output data in tiles, as shown in Figure~\ref{fig:tiling}. 
Each MSA can be seen as multiple parallel matrix multiplications. The accelerator is designed to process parallel computations within each head.
This input channel splitting is also done for fully connected (FC) layers, each containing one matrix multiplication for compatibility, and the results need to be accumulated from all the input channels in all the heads.  
Furthermore, the computing engine exploits fine-grained data and operation parallelisms and can process $T_m \cdot T_n$ multiply-accumulate (MAC) operations in parallel.

We model the inference latency of our hardware design based on the number of clock cycles. For a layer $i$ in ViT, since data packing can be used to transfer multiple (i.e., $D_{act}$ or $D_{wgt}$) values at the same time in each AXI port,
the clock cycles for loading one input/weight tile and storing one output tile, are calculated as: 
\begin{equation}
\small
\begin{aligned}
L_{\mathrm{in}} &=  \left\lceil \frac{T_n}{D_{act}} \right\rceil \cdot \left\lceil \frac{F}{A_{\mathrm{in}}} \right\rceil, \\
L_{\mathrm{wgt}} &=   \left\lceil \frac{T_n}{D_{wgt}} \right\rceil \cdot \left\lceil \frac{T_m}{A_{\mathrm{wgt}}} \right\rceil, \\
L_{\mathrm{out}} &=  \left\lceil \frac{T_m}{D_{act}} \right\rceil \cdot \left\lceil \frac{F}{A_{\mathrm{out}}} \right\rceil,
\end{aligned}
\end{equation}
where $L_{\mathrm{in}}$, $L_{\mathrm{wgt}}$ and $L_{\mathrm{out}}$ indicate the number of the clock cycles of input, weight, and output transfer for one corresponding tile, respectively.

The clock cycle number to compute one tile is 
\begin{equation}
\small
L_{\mathrm{cmpt}} = \max \big\{\left\lceil \frac{F}{P_F} \right\rceil, \frac{T_n \cdot T_m \cdot F}{N_{tot}} \big\},
\end{equation}
where the first term is calculated by the latency model. The total number of multiplications needed to compute one tile of matrix multiplication is $T_n \cdot T_m \cdot F$.
Each tile has two levels of parallelism: 
one is along the $T_n$ and $T_m$ dimension and the parallel factor is $T_n \cdot T_m$ (i.e., fully parallel); and the other is along the $F$ dimension and the parallel factor is $P_F$. Therefore, the first term is calculated as $\left\lceil \frac{F}{P_F} \right\rceil$. The second term is limited by the resource constraint, where we can support at most $N_{tot}$ parallel multipliers (Eq.~\ref{eqn:max}) due to the resource constraint.

With the double buffers overlapping the data loading and computation of the tiles, the overall clock cycle number for processing one tile is $L_1 = \max\{L_{\mathrm{in}}, L_{\mathrm{wgt}}, L_{\mathrm{cmpt}}\}$.

\begin{table*}[!t]
\centering
\caption{Comparison of representative ViT-based works with our proposed \M.}
\resizebox{2.0 \columnwidth}{!}
{
\begin{tabular}{lcccccccc}
\toprule
\multirow{2}*{Model} &  Quantization  & \#Params  & Model Size  & MACs     & \multicolumn{2}{c}{MAC Bit-width}  & Equivalent & Top-1 Accuracy\\
   & Mixed Type   & $(M)$ & $(MB)$   &  $(G)$   & Weight & Activation & BOPs $(G)$ & $(\% )$  \\
\midrule
DeiT-T~\cite{Touvron2021TrainingDI}  & - & 5.7 & 22.8 & 1.3  & 32 & 32  & $1.3\times10^3$ & 72.2  \\
T2T-T~\cite{Yuan_2021_ICCV}   & - & 4.3 & 17.2 & 1.1   & 32 & 32 & $1.1\times10^3$ & 71.7  \\
PiT-T~\cite{heo2021rethinking}   & - & 10.6 & 42.4 & 1.4   & 32 & 32 & $1.4\times10^3$ & 72.4  \\
LocalViT-T~\cite{li2021localvit} & - & 5.9 & 23.6 & 1.3 & 32 & 32 & $1.3\times10^3$ & 74.8 \\
FQ-ViT~\cite{lin2022fq} & Model-wise & 5.7 & 5.7 & 1.3 & 8 & 8 & 80.6 & 71.6 \\
Q-ViT~\cite{li2022q} & Head-wise & 5.7 & - & - & 4-8 & 4-8 & - & 72.8 \\
\textbf{QUASAR-S (ours)} & Row-wise  & 5.9 & \textbf{4.1} & 1.4  & 4 \& 8 & 6  & \textbf{45.6} & \textbf{74.9} \\
\hline
PVT~\cite{wang2021pyramid}   & - & 24.5 & 98.0 & 3.8 & 32 & 32 & $3.9\times10^3$ & 79.8 \\
DeiT-S~\cite{Touvron2021TrainingDI}  & - & 22.9 & 91.6 & 4.6  & 32 & 32  & $4.8\times10^3$ & 79.9  \\
Swin-T~\cite{liu2021swin} & - & 28 & 112.0 & 4.5 & 32 & 32 & $4.7\times10^3$ & 81.2 \\
BossNAS~\cite{li2021bossnas} & - & - & -& 3.4 & 32 & 32 & $3.5\times10^3$ & 80.5 \\
PTQ~\cite{liu2021post} & Layer-wise & 22.9 & 16.6 & 4.6 & 6-10 & 6-10 & - & 75.1\\
FQ-ViT~\cite{lin2022fq} & Model-wise & 22.9 & 22.9 & 4.6 & 8 & 8  & 294.4 & 79.1 \\
Q-ViT~\cite{li2022q} & Head-wise & 22.9 & - & - & 4-8 & 4-8 & - & 80.1 \\ 
\textbf{QUASAR-L1 (ours)}   & Row-wise  & 14.7 & \textbf{9.8} & 3.2 & 4 \& 8 & 6 & \textbf{103.8} & \textbf{78.6} \\
\textbf{QUASAR-L2 (ours)}   & Row-wise  & 22.6 & \textbf{15.8} & 4.8 & 4 \& 8 & 6 & \textbf{163.2} & \textbf{80.4} \\
\bottomrule
\end{tabular}
}
\label{table:accuracy}
\end{table*}

To obtain the accumulation of output results, this process
is performed multiple times (i.e., processing $\left\lceil \frac{N}{T_n} \right\rceil$ input tiles). The clock cycle number for
calculating the whole output tile is $L_2 = \max \big\{L_1 \cdot \left\lceil \frac{N}{T_n} \right\rceil + L_{\mathrm{cmpt}}, L_{\mathrm{out}} \big\}$. 

Since there are $\left\lceil \frac{M}{T_m} \right\rceil$ number of output tiles, the total number of clock cycles for a ViT layer $i$ is described by

\begin{equation}
\small
L_{\mathrm{tot}}^i = \left\lceil \frac{M}{T_m} \right\rceil \cdot L_2 + L_{\mathrm{out}}.
\label{clock_cycle}
\end{equation}

Under a working frequency $f$, the $\mathrm{FPS}$ is calculated as:
\begin{equation}
\small
FPS = \frac{f}{\sum\limits_i L_{\mathrm{tot}}^i}.
\label{fps}
\end{equation}

\section{Experimental Results}\label{sec:results}

\subsection{Experimental Setup}
\label{sec:setting}

Our supernet training process takes 700 epochs with a batch size of 2048. The learning rate is set to $5 \times 10^{-4}$ initially and decayed with a cosine annealing schedule. The AdamW~\cite{loshchilov2018decoupled} optimizer is used with the epsilon value of $1e^{-8}$ and weight decay of 0.05. Additional training optimizations, such as warmup and label smoothing are performed during training. The number of warmup epochs and label smoothing factor are set as 20 and 0.1, respectively. After supernet training, we perform the hardware-oriented evolution search, without subnet retraining. 

Our training is conducted on 8 NVIDIA Ampere A100 GPUs with CUDA 11.0 and PyTorch 1.7 frameworks on the Ubuntu operating system.
To test the effectiveness of our framework, we also implement~\M~framework on the Xilinx ZCU102 embedded FPGA platform with quad-core ARM Cortex-A53 and XCZU9EG FPGA chip. The FPGA working frequency is set to 150 MHz for all the designs implemented via Xilinx Vitis and Vitis HLS 2020.1.

\begin{table}[h] 
\centering
\caption{\rev{Comparison between different settings of QUASAR-Small with and without knowledge distillation (KD) and supernet layer scaling (SLS) on ImageNet dataset. }
}
\scalebox{0.9}{
\begin{tabular}{c|cc|ccc}
\toprule
\multirow{2}*{Model} & \multicolumn{2}{c|}{Setting} & \#Params & Quantization & Top-1 Acc.   \\
 & KD & SLS & (M) & Scheme & (\%) \\
\midrule
DeiT-T~\cite{Touvron2021TrainingDI} & No & No & 5.7  & W32A32  &  72.2 \\
DeiT-T (quant) & No & No & 5.7 & W8A8  &  71.5 \\
Ours & No & No & 5.9 & W8A8 & 74.1 \\
Ours &Yes & No & 5.9 & W8A8 & 75.6 \\
Ours &No & Yes & 5.9 & W8A8 & 74.9  \\
Ours &Yes & Yes & 5.9 & W8A8 & 76.1 \\
\bottomrule
\end{tabular}
}
\label{table:ablation}
\end{table}

\subsection{Accuracy Results}

Here we analyze the accuracy results of our~\M~framework. The weight precision is mixed with 4-bit and 8-bit, as mentioned earlier, and the activation bit-width is determined by the hardware feature. 
Without loss of generality, we use activation of 8-bit to evaluate the accuracy in the ablation study of knowledge distillation and supernet layer
scaling.

\subsubsection{Ablation Study of Knowledge Distillation and Supernet Layer Scaling}

To evaluate the compatibility of knowledge distillation and our proposed SLS, we conduct an ablation study on both of them. Without loss of generality and to prevent interference from different model sizes and different quantization mixed ratios from the searched subnet, we unify the search constraint with pure W8A8 (8-bit for both weight and activation) quantization implementation.

As shown in Table~\ref{table:ablation}, we conduct four different settings of \M~with the unified model size and quantization scheme. The accuracy of the four cases is 74.1\% (w/o distillation and SLS), 75.6\% (only distillation), 74.9\% (only SLS), and 76.1\% (with both of them), respectively. Knowledge distillation and SLS strategies are orthogonal to each other, and both improve the model accuracy. Using them together provides a better result. Given the observed effectiveness of our proposed SLS strategy and the seamless compatibility of our framework with knowledge distillation, we opt to incorporate both strategies in our following experiments. \rev{Here we also quantize the baseline DeiT-T~\cite{Touvron2021TrainingDI} and compare it with our method without SLS and distillation. Even without SLS and distillation, our quantization NAS approach achieves a much better model accuracy than the full-precision and the quantized (W8A8) DeiT-T models.}

\subsubsection{\rev{Ablation Study of Mixed-Ratio and Quantization Scheme}}

\rev{To assess the efficacy of our row-wise flexible mixed-precision quantization scheme, we conducted an ablation study examining both the quantization scheme itself and the 8-bit mixed ratio, as outlined in Table~\ref{tab:scheme}. Since~\M~ automatically searches the mixed ratios, here we pick up the best model from the search stage for different mixed ratios and compare them with the counterparts under the fixed row-wise mixed quantization scheme. The results indicate a consistent improvement in accuracy across different 8-bit mixed-ratio levels with our flexible mixed scheme, underscoring the efficiency of our proposed quantization scheme.}

\begin{table}[h] 
\centering
\caption{\rev{Comparison between different settings of 8-bit quantization mixed-ratio and quantization schemes.}}
\scalebox{0.9}{
\begin{tabular}{c|ccc}
\toprule
 Scheme   & Model Size (MB) & 8-bit mixed-ratio (\%)   & Acc. (\%)  \\ 
\midrule
Fixed row-wise   & 3.6    & 23  & 73.5     \\  
Flexible row-wise  & 3.6  & 23  & 74.2     \\   
Fixed row-wise   & 4.1   & 39  & 74.1     \\ 
Flexible row-wise  & 4.1  & 39  & 74.9     \\ 
\bottomrule
\end{tabular}}
\label{tab:scheme}
\end{table}

\subsubsection{Overall Accuracy Results}

Table~\ref{table:accuracy} compares representative ViT-based works with our proposed \M. 
Since many ViT-based works do not incorporate model quantization, we also consider the bit-width in the model size and the equivalent number of total bit operations (BOPs).

As shown in Table~\ref{table:accuracy}, we present three Quasar models for different accuracy levels: QUASAR-S searched within the QUASAR-Small supernet, and QUASAR-L1, QUASAR-L2 searched within the QUASAR-Large supernet. 

Our QUASAR-S achieves 74.9\% top-1 accuracy with only 4.1 MB model size and 1.4 GMACs, 
Compared with the representative ViT-based model LocalViT-T~\cite{li2021localvit} under a similar accuracy, our model size is only 17.4\% of that in LocalViT-T; although the GMACs numbers are similar, our MAC unit is much more hardware efficient as it is for a 4-bit/8-bit weight and a 6-bit activation instead of a 32-bit floating-point MAC unit in LocalViT-T.
For a higher model accuracy level, our QUASAR-L1 and QUASAR-L2 achieve 78.5\% and 80.4\% top-1 accuracy, respectively. Among them, QUASAR-L2 only has a 15.8 MB model size with a computation volume of 4.8 GMACs, which obtains the smallest BOPs with a similar level of accuracy compared with other baselines. Specifically, compared with PTQ~\cite{liu2021post} (16.6 MB, top-1 75.1\%), QUASAR-L2 achieves a similar model size and GMACs with 5.2\% higher accuracy. Compared with ViT NAS framework BossNAS~\cite{li2021bossnas}, we additionally achieve low-bit quantization with a much smaller BOPS and similar accuracy.
Compared with quantization-aware training framework Q-ViT~\cite{li2022q} using multiple quantization bit-widths in the range of 4 to 8, which incurs inefficiency and hardware under-utilization, our results show better accuracy with a more unified and hardware-friendly quantization scheme.

\begin{table}[!tbp]
\tabcolsep 4pt
\caption{Comparisons of FPGA implementations for ViTs on ImageNet, including DeiT-S and Auto-ViT-Acc from \cite{liauto}, and our \M{}, all running at 150MHz on the same AMD/Xilinx ZCU102 embedded FPGA platform.}

\small
\centering
\scalebox{1.0}{
\begin{tabular}{c|c|c|c|c|c}
\toprule
\multirow{2}{*}{} & \multirow{2}{*}{DeiT-S} & Auto-ViT &  \multicolumn{3}{c}{\M{}} \\
 &  & -Acc & L2 & L1 & S \\
\midrule
\multirow{2}{*}{Quant.}  & \multirow{2}{*}{No} & Mixed & Mixed & Mixed & Mixed \\
 &  & Scheme & Precision & Precision & Precision \\
\hline
Weight      & 32     & 4 \& 8  & 4 \& 8 & 4 \& 8  & 4 \& 8  \\ \hline
Act.        & 32     & 8       & 6      & 6  & 6       \\ \hline
Top-1 Acc.  & 79.9   & 78.7    & 80.4  & 78.6  & 74.9    \\ \hline
kLUT        & 47\%   & 67\%    & 66\%  & 66\%  & 65\%    \\
FF          & -      & -       & 31\%  & 31\% & 30\%    \\
DSP         & 69\%   & 62\%    & 69\%  &  69\% & 69\%    \\
BRAM        & -      & -       & 44\%  &  44\% & 44\%    \\
\bottomrule
\end{tabular}
}
\label{tab:hardware}
\end{table}

\subsection{Comparison of Hardware Results on FPGA}

We implement a proof-of-concept hardware accelerator for our \M~on the AMD/Xilinx ZCU102 embedded FPGA platform.
We also compare our results to Auto-ViT-Acc~\cite{liauto}, the state-of-the-art FPGA accelerator for ViT with mixed-scheme quantization (without NAS). We retrieve the hardware results for Auto-ViT-Acc (which is quantized from DeiT-S) and the original DeiT-S on the same Xilinx ZCU102 FPGA platform from~\cite{liauto}.

\begin{figure}[htbp]
  \centering
  \includegraphics[width=1.0\linewidth]{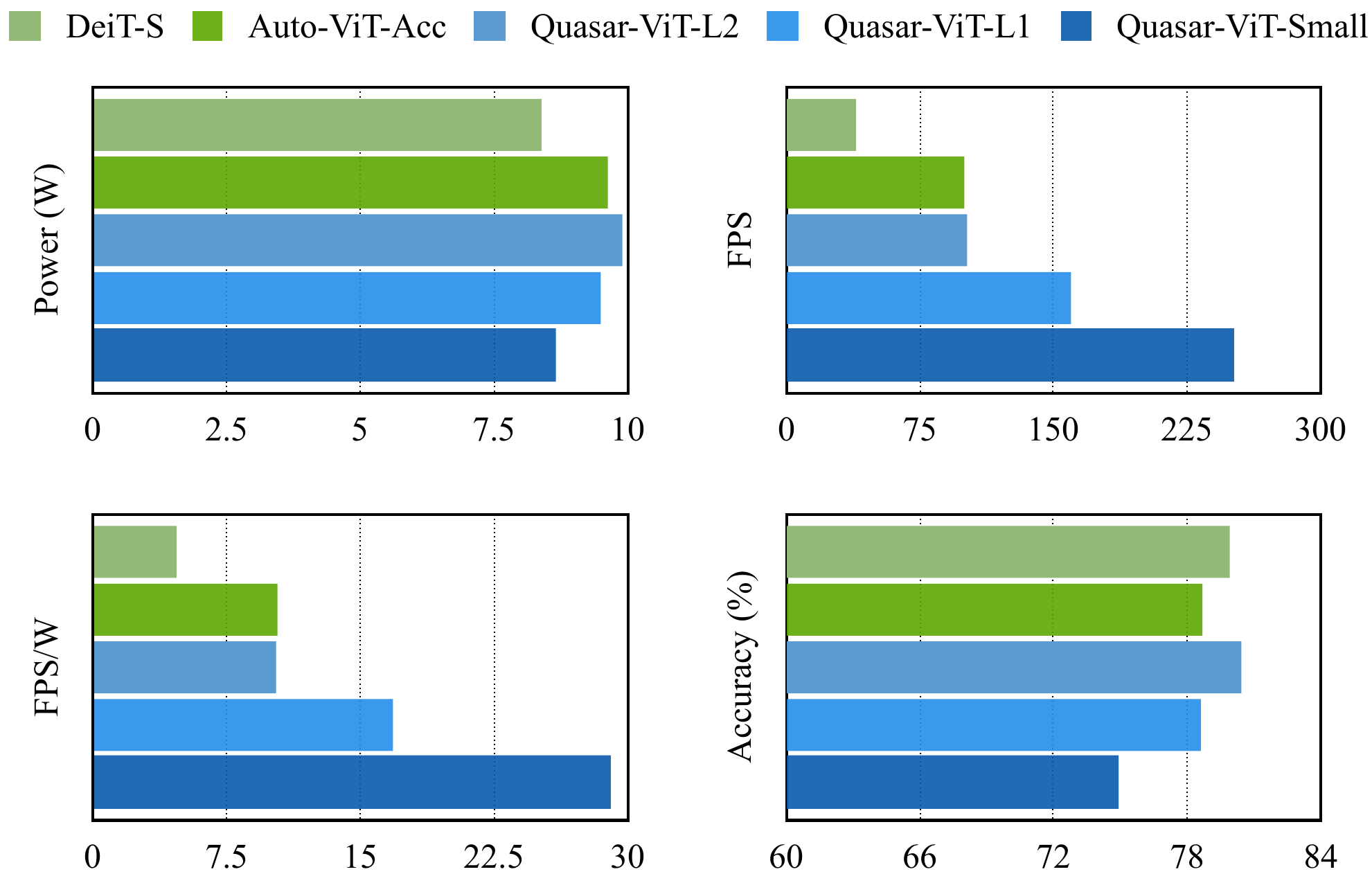}
  \caption{Comparisons between DeiT-S, Auto-ViT-Acc from~\cite{liauto}, and our \M{}.
  }
  \label{fig:comp}
\end{figure}

As shown in Table~\ref{tab:hardware} and Figure~\ref{fig:comp}, our approach consistently outperforms the previous work. Specifically, compared with DeiT-S~\cite{Touvron2021TrainingDI}, our QUASAR-L2 achieves 2.6$\times$ higher inference frames per second (FPS) with 0.5\% better accuracy. Compared with Auto-ViT-Acc~\cite{liauto}, our QUASAR-L1 achieves 1.6$\times$ higher FPS (159.6) with a similar model accuracy level, and our QUASAR-L2 achieves a similar level of FPS with 1.7\% better top-1 accuracy.

The improvement in model accuracy and inference performance within our framework is attributed to two key factors. Firstly, our approach involves the training and search for a customized network architecture, specifically tailored for both the mixed-precision quantization schemes and the targeted inference latency. This strategy enhances adaptability and efficiency, surpassing the achievements of previous methodologies.

Secondly, our novel supernet training algorithm, coupled with the proposed hybrid DSP packing design, allows for distinct quantization mixed ratios across various model layers. This fine-grained model achieves better flexibility than the previous approaches, unleashing the full potential of mixed-precision quantization.

\rev{With regard to the efficiency of our hardware accelerator design, the performance is mainly limited by DSP, LUT, and off-chip memory bandwidth. On par with Auto-ViT-Acc~\cite{liauto}, our design achieves 150MHz frequency with about 66\% usage of LUTs and 69\% DSPs without timing violations. 
Note a typical FPGA design usually utilizes approximately 60\% to 70\% of the available FPGA resources; otherwise, it may fail during the placement and routing phase due to congestion or result in a lower operating frequency. 
Without considering the timing violation, the maximum theoretical expected performance is based on the 100\% utilization ratio for both DSP, LUTs, and bandwidth, which can achieve about 1.47x of our reached FPS for the same model.}

\subsection{Other Transformer-based Model Accuracy Results}

\rev{To demonstrate the scalability and versatility of our methods, we applied them across various datasets and applications, notably deploying them on a large language model (LLM). This choice is motivated by two key factors. Firstly, LLM shares a transformer-based architecture similar to that of Vision Transformer (ViT), aligning with the framework that we propose. Secondly, LLM is frequently integrated with ViT in text-to-image/video applications, making it an ideal candidate to showcase the scalability of our approach across both models and its potential for real-world applications.}

\rev{Our comparative analysis, presented in Table~\ref{tab:datallm}, utilizes the renowned LLM model, LLaMA, as the foundation for our supernet. We juxtapose our optimized results with those of LLaMA-7B~\cite{touvron2023llama} on the commonly used WikiText-2 dataset for LLMs, with perplexity score (PPL) serving as the evaluation metric, where lower scores indicate superior performance. According to the comparison results, our method shows a constant pattern, achieving a similar level of PPL with a much smaller model size.}

\begin{table}[h] 
\centering
\caption{\rev{Result comparison on large language model.}}
\scalebox{0.9}{
\begin{tabular}{c|cccc}
\toprule
 Model & Model Size (GB) & W \# Bits  & A \# Bits &  PPL  \\ 
\midrule
LLaMA-7B~\cite{touvron2023llama} & 26.8 & FP32 & FP32 & 5.68\\
Ours & 6.7 & INT8 & INT8  & 5.73\\
Ours & 4.8 & INT4 \& 8    & INT8 &  5.91 \\
\bottomrule
\end{tabular}}
\label{tab:datallm}
\end{table}

\subsection{Training Cost Comparison}

\rev{Prior co-design frameworks, such as APQ~\cite{wang2020apq}, have also delved into the integration of neural architecture search (NAS) and quantization techniques. Please note that APQ is based on the convolutional neural network (CNN) and BitFusion platform~\cite{wang2020apq}. To the best of our knowledge, we compare our Quasar-ViT models (both small and large variants) and the APQ result. As detailed in Table~\ref{tab:cost}, our approach demonstrates superior FPS performance while maintaining comparable or even higher model accuracy, achieved at a reduced training cost. Compared with the 2,400 GPU hours training cost of APQ~\cite{wang2020apq}, our approach only consumes 768 and 1,344 GPU hours for the small and large versions of~\M, respectively. Our training setting has been illustrated in Section~\ref{sec:setting}.}

\begin{table}[h] 
\centering
\caption{\rev{Training cost and accuracy comparison with other NAS and quantization co-design.}}
\scalebox{0.9}{
\begin{tabular}{c|ccc}
\toprule
 Method & Training cost (GPU hours) & FPS  & Acc.(\%)  \\ 
\midrule
APQ~\cite{wang2020apq}   & 2400  &  82.2  &   75.1     \\
QUASAR-S  & 768 & 101.5   &   74.9 \\
QUASAR-L  & 1344  & 251.6 &   80.4  \\
\bottomrule
\end{tabular}}
\label{tab:cost}
\end{table}

\section{Conclusion}\label{sec:concl}

In this work, we propose~\M, a hardware-oriented quantization-aware network architecture search framework to enable efficient ViT deployment on resource-constrained edge devices.
First, we proposed hardware-friendly quantization techniques including flexible row-wise mixed-precision quantization scheme and intra-layer mixed-precision weight entanglement in architecture search towards high accuracy and low training cost for efficient implementation.
Second, we propose 4-bit weight atomic computation and hybrid signed/unsigned DSP packing for FPGA implementation, then incorporate latency/resource modeling to enable the hardware-oriented architecture search.
Third, we extend the supernet layer scaling technique to further improve the training convergence and supernet accuracy. We also demonstrate the compatibility of our proposed framework with knowledge distillation during supernet training. Finally, we developed an efficient hardware-oriented search algorithm---integrated with hardware latency and resource modeling---to search the efficient subnet with high accuracy under a given inference latency target and implemented the searched model on real FPGA hardware for validation.
From the experiment evaluation results, our approach achieves 101.5, 159.6, and 251.6 FPS on the AMD/Xilinx ZCU102 FPGA board with 80.4\%, 78.6\%, and 74.9\% top-1 accuracy for ImageNet, respectively, consistently outperforming prior works.

\begin{acks}
This work was supported in part by NSERC Discovery Grant RGPIN-2019-04613, DGECR-2019-00120, Alliance Grant ALLRP-552042-2020; CFI John R. Evans Leaders Fund; BC Knowledge Development Fund; Army Research Office/Army Research Laboratory via grant W911-NF-20-1-0167 to Northeastern University; National Science Foundation CCF-1937500, CNS-1909172, and IIS-2310254. No external funding support to CD and MS for this project.
% SRC Artificial Intelligence Hardware program (CD) 3191.001.
\end{acks}

% \input{Sections/appendix.tex}

% \begin{acks}
% This study was supported by JST PRESTO Program (Grant No. JPMJPR19M7), FOREST Program (Grant Number JPMJFR226W, Japan), and the NSF Expedition program CCF-2124453, NSF CCF-2008514.
% \end{acks}

%%
%% The next two lines define the bibliography style to be used, and
%% the bibliography file.
\bibliographystyle{ACM-Reference-Format}
% \bibliography{sample-base}
\bibliography{references}

%%% -*-BibTeX-*-
%%% Do NOT edit. File created by BibTeX with style
%%% ACM-Reference-Format-Journals [18-Jan-2012].

\begin{thebibliography}{70}

%%% ====================================================================
%%% NOTE TO THE USER: you can override these defaults by providing
%%% customized versions of any of these macros before the \bibliography
%%% command.  Each of them MUST provide its own final punctuation,
%%% except for \shownote{}, \showDOI{}, and \showURL{}.  The latter two
%%% do not use final punctuation, in order to avoid confusing it with
%%% the Web address.
%%%
%%% To suppress output of a particular field, define its macro to expand
%%% to an empty string, or better, \unskip, like this:
%%%
%%% \newcommand{\showDOI}[1]{\unskip}   % LaTeX syntax
%%%
%%% \def \showDOI #1{\unskip}           % plain TeX syntax
%%%
%%% ====================================================================

\ifx \showCODEN    \undefined \def \showCODEN     #1{\unskip}     \fi
\ifx \showDOI      \undefined \def \showDOI       #1{#1}\fi
\ifx \showISBNx    \undefined \def \showISBNx     #1{\unskip}     \fi
\ifx \showISBNxiii \undefined \def \showISBNxiii  #1{\unskip}     \fi
\ifx \showISSN     \undefined \def \showISSN      #1{\unskip}     \fi
\ifx \showLCCN     \undefined \def \showLCCN      #1{\unskip}     \fi
\ifx \shownote     \undefined \def \shownote      #1{#1}          \fi
\ifx \showarticletitle \undefined \def \showarticletitle #1{#1}   \fi
\ifx \showURL      \undefined \def \showURL       {\relax}        \fi
% The following commands are used for tagged output and should be
% invisible to TeX
\providecommand\bibfield[2]{#2}
\providecommand\bibinfo[2]{#2}
\providecommand\natexlab[1]{#1}
\providecommand\showeprint[2][]{arXiv:#2}

\bibitem[Bai et~al\mbox{.}(2021)]%
        {bai2020binarybert}
\bibfield{author}{\bibinfo{person}{Haoli Bai}, \bibinfo{person}{Wei Zhang}, \bibinfo{person}{Lu Hou}, \bibinfo{person}{Lifeng Shang}, \bibinfo{person}{Jing Jin}, \bibinfo{person}{Xin Jiang}, \bibinfo{person}{Qun Liu}, \bibinfo{person}{Michael Lyu}, {and} \bibinfo{person}{Irwin King}.} \bibinfo{year}{2021}\natexlab{}.
\newblock \showarticletitle{{B}inary{BERT}: Pushing the Limit of {BERT} Quantization}. In \bibinfo{booktitle}{\emph{Proceedings of the 59th Annual Meeting of the Association for Computational Linguistics and the 11th International Joint Conference on Natural Language Processing (Volume 1: Long Papers)}}.
\newblock


\bibitem[Baker et~al\mbox{.}(2017)]%
        {baker2016designing}
\bibfield{author}{\bibinfo{person}{Bowen Baker}, \bibinfo{person}{Otkrist Gupta}, \bibinfo{person}{Nikhil Naik}, {and} \bibinfo{person}{Ramesh Raskar}.} \bibinfo{year}{2017}\natexlab{}.
\newblock \showarticletitle{Designing Neural Network Architectures using Reinforcement Learning}. In \bibinfo{booktitle}{\emph{International Conference on Learning Representations (ICLR)}}.
\newblock


\bibitem[Bender et~al\mbox{.}(2018)]%
        {bender2018understanding}
\bibfield{author}{\bibinfo{person}{Gabriel Bender}, \bibinfo{person}{Pieter-Jan Kindermans}, \bibinfo{person}{Barret Zoph}, \bibinfo{person}{Vijay Vasudevan}, {and} \bibinfo{person}{Quoc Le}.} \bibinfo{year}{2018}\natexlab{}.
\newblock \showarticletitle{Understanding and simplifying one-shot architecture search}. In \bibinfo{booktitle}{\emph{Proceedings of the International Conference on Machine Learning (ICML)}}. \bibinfo{pages}{550--559}.
\newblock


\bibitem[Cai et~al\mbox{.}(2018)]%
        {cai2017efficient}
\bibfield{author}{\bibinfo{person}{Han Cai}, \bibinfo{person}{Tianyao Chen}, \bibinfo{person}{Weinan Zhang}, \bibinfo{person}{Yong Yu}, {and} \bibinfo{person}{Jun Wang}.} \bibinfo{year}{2018}\natexlab{}.
\newblock \showarticletitle{Efficient architecture search by network transformation}. In \bibinfo{booktitle}{\emph{Proceedings of the AAAI Conference on Artificial Intelligence}}, Vol.~\bibinfo{volume}{32}.
\newblock


\bibitem[Chang et~al\mbox{.}(2021)]%
        {chang2021mix}
\bibfield{author}{\bibinfo{person}{Sung-En Chang}, \bibinfo{person}{Yanyu Li}, \bibinfo{person}{Mengshu Sun}, \bibinfo{person}{Runbin Shi}, \bibinfo{person}{Hayden K-H So}, \bibinfo{person}{Xuehai Qian}, \bibinfo{person}{Yanzhi Wang}, {and} \bibinfo{person}{Xue Lin}.} \bibinfo{year}{2021}\natexlab{}.
\newblock \showarticletitle{Mix and Match: A novel FPGA-centric deep neural network quantization framework}. In \bibinfo{booktitle}{\emph{2021 IEEE International Symposium on High-Performance Computer Architecture (HPCA)}}. \bibinfo{pages}{208--220}.
\newblock


\bibitem[Chen et~al\mbox{.}(2021)]%
        {chen2021autoformer}
\bibfield{author}{\bibinfo{person}{Minghao Chen}, \bibinfo{person}{Houwen Peng}, \bibinfo{person}{Jianlong Fu}, {and} \bibinfo{person}{Haibin Ling}.} \bibinfo{year}{2021}\natexlab{}.
\newblock \showarticletitle{Autoformer: Searching transformers for visual recognition}. In \bibinfo{booktitle}{\emph{Proceedings of the IEEE/CVF International Conference on Computer Vision (ICCV)}}. \bibinfo{pages}{12270--12280}.
\newblock


\bibitem[Choi et~al\mbox{.}(2018)]%
        {choi2018pact}
\bibfield{author}{\bibinfo{person}{Jungwook Choi}, \bibinfo{person}{Zhuo Wang}, \bibinfo{person}{Swagath Venkataramani}, \bibinfo{person}{Pierce I-Jen Chuang}, \bibinfo{person}{Vijayalakshmi Srinivasan}, {and} \bibinfo{person}{Kailash Gopalakrishnan}.} \bibinfo{year}{2018}\natexlab{}.
\newblock \showarticletitle{Pact: Parameterized clipping activation for quantized neural networks}.
\newblock \bibinfo{journal}{\emph{arXiv:1805.06085}} (\bibinfo{year}{2018}).
\newblock


\bibitem[Courbariaux et~al\mbox{.}(2015)]%
        {courbariaux2015binaryconnect}
\bibfield{author}{\bibinfo{person}{Matthieu Courbariaux}, \bibinfo{person}{Yoshua Bengio}, {and} \bibinfo{person}{Jean-Pierre David}.} \bibinfo{year}{2015}\natexlab{}.
\newblock \showarticletitle{Binaryconnect: Training deep neural networks with binary weights during propagations}. In \bibinfo{booktitle}{\emph{Advances in Neural Information Processing Systems (NeurIPS)}}. \bibinfo{pages}{3123--3131}.
\newblock


\bibitem[Dong et~al\mbox{.}(2019)]%
        {dong2019hawq}
\bibfield{author}{\bibinfo{person}{Zhen Dong}, \bibinfo{person}{Zhewei Yao}, \bibinfo{person}{Amir Gholami}, \bibinfo{person}{Michael~W Mahoney}, {and} \bibinfo{person}{Kurt Keutzer}.} \bibinfo{year}{2019}\natexlab{}.
\newblock \showarticletitle{Hawq: Hessian aware quantization of neural networks with mixed-precision}. In \bibinfo{booktitle}{\emph{Proceedings of the IEEE/CVF International Conference on Computer Vision (ICCV)}}.
\newblock


\bibitem[Dosovitskiy et~al\mbox{.}(2021)]%
        {dosovitskiy2020image}
\bibfield{author}{\bibinfo{person}{Alexey Dosovitskiy}, \bibinfo{person}{Lucas Beyer}, \bibinfo{person}{Alexander Kolesnikov}, \bibinfo{person}{Dirk Weissenborn}, \bibinfo{person}{Xiaohua Zhai}, \bibinfo{person}{Thomas Unterthiner}, \bibinfo{person}{Mostafa Dehghani}, \bibinfo{person}{Matthias Minderer}, \bibinfo{person}{Georg Heigold}, \bibinfo{person}{Sylvain Gelly}, \bibinfo{person}{Jakob Uszkoreit}, {and} \bibinfo{person}{Neil Houlsby}.} \bibinfo{year}{2021}\natexlab{}.
\newblock \showarticletitle{An Image is Worth 16x16 Words: Transformers for Image Recognition at Scale}. In \bibinfo{booktitle}{\emph{International Conference on Learning Representations (ICLR)}}.
\newblock
\urldef\tempurl%
\url{https://openreview.net/forum?id=YicbFdNTTy}
\showURL{%
\tempurl}


\bibitem[Guo et~al\mbox{.}(2020)]%
        {guo2020single}
\bibfield{author}{\bibinfo{person}{Zichao Guo}, \bibinfo{person}{Xiangyu Zhang}, \bibinfo{person}{Haoyuan Mu}, \bibinfo{person}{Wen Heng}, \bibinfo{person}{Zechun Liu}, \bibinfo{person}{Yichen Wei}, {and} \bibinfo{person}{Jian Sun}.} \bibinfo{year}{2020}\natexlab{}.
\newblock \showarticletitle{Single path one-shot neural architecture search with uniform sampling}. In \bibinfo{booktitle}{\emph{Proceedings of the European Conference on Computer Vision (ECCV)}}. Springer, \bibinfo{pages}{544--560}.
\newblock


\bibitem[He et~al\mbox{.}(2016)]%
        {he2016deep}
\bibfield{author}{\bibinfo{person}{Kaiming He}, \bibinfo{person}{Xiangyu Zhang}, \bibinfo{person}{Shaoqing Ren}, {and} \bibinfo{person}{Jian Sun}.} \bibinfo{year}{2016}\natexlab{}.
\newblock \showarticletitle{Deep residual learning for image recognition}. In \bibinfo{booktitle}{\emph{Proceedings of the IEEE/CVF Conference on Computer Vision and Pattern Recognition (CVPR)}}. \bibinfo{pages}{770--778}.
\newblock


\bibitem[He and Fan(2019)]%
        {he2019simultaneously}
\bibfield{author}{\bibinfo{person}{Zhezhi He} {and} \bibinfo{person}{Deliang Fan}.} \bibinfo{year}{2019}\natexlab{}.
\newblock \showarticletitle{Simultaneously optimizing weight and quantizer of ternary neural network using truncated gaussian approximation}. In \bibinfo{booktitle}{\emph{Proceedings of the IEEE/CVF Conference on Computer Vision and Pattern Recognition (CVPR)}}.
\newblock


\bibitem[Heo et~al\mbox{.}(2021)]%
        {heo2021rethinking}
\bibfield{author}{\bibinfo{person}{Byeongho Heo}, \bibinfo{person}{Sangdoo Yun}, \bibinfo{person}{Dongyoon Han}, \bibinfo{person}{Sanghyuk Chun}, \bibinfo{person}{Junsuk Choe}, {and} \bibinfo{person}{Seong~Joon Oh}.} \bibinfo{year}{2021}\natexlab{}.
\newblock \showarticletitle{Rethinking spatial dimensions of vision transformers}.
\newblock \bibinfo{journal}{\emph{arXiv:2103.16302}} (\bibinfo{year}{2021}).
\newblock


\bibitem[Hinton et~al\mbox{.}(2015)]%
        {hinton2015distilling}
\bibfield{author}{\bibinfo{person}{Geoffrey Hinton}, \bibinfo{person}{Oriol Vinyals}, {and} \bibinfo{person}{Jeff Dean}.} \bibinfo{year}{2015}\natexlab{}.
\newblock \showarticletitle{Distilling the knowledge in a neural network}.
\newblock \bibinfo{journal}{\emph{arXiv:1503.02531}} (\bibinfo{year}{2015}).
\newblock


\bibitem[Intel(2024)]%
        {intel_cpu}
\bibfield{author}{\bibinfo{person}{Intel}.} \bibinfo{year}{2024}\natexlab{}.
\newblock \bibinfo{title}{Intel® Xeon® Silver 4214 Processor}.
\newblock \bibinfo{howpublished}{\url{https://ark.intel.com/content/www/us/en/ark/products/193385/intel-xeon-silver-4214-processor-16-5m-cache-2-20-ghz.html}}.
\newblock
\newblock
\shownote{Last accessed Jan. 18, 2024}.


\bibitem[Leng et~al\mbox{.}(2018)]%
        {leng2018extremely}
\bibfield{author}{\bibinfo{person}{Cong Leng}, \bibinfo{person}{Zesheng Dou}, \bibinfo{person}{Hao Li}, \bibinfo{person}{Shenghuo Zhu}, {and} \bibinfo{person}{Rong Jin}.} \bibinfo{year}{2018}\natexlab{}.
\newblock \showarticletitle{Extremely low bit neural network: Squeeze the last bit out with admm}. In \bibinfo{booktitle}{\emph{Proceedings of the AAAI Conference on Artificial Intelligence}}.
\newblock


\bibitem[Li et~al\mbox{.}(2021a)]%
        {li2021bossnas}
\bibfield{author}{\bibinfo{person}{Changlin Li}, \bibinfo{person}{Tao Tang}, \bibinfo{person}{Guangrun Wang}, \bibinfo{person}{Jiefeng Peng}, \bibinfo{person}{Bing Wang}, \bibinfo{person}{Xiaodan Liang}, {and} \bibinfo{person}{Xiaojun Chang}.} \bibinfo{year}{2021}\natexlab{a}.
\newblock \showarticletitle{Bossnas: Exploring hybrid cnn-transformers with block-wisely self-supervised neural architecture search}. In \bibinfo{booktitle}{\emph{Proceedings of the IEEE/CVF International Conference on Computer Vision (ICCV)}}. \bibinfo{pages}{12281--12291}.
\newblock


\bibitem[Li et~al\mbox{.}(2020)]%
        {li2020additive}
\bibfield{author}{\bibinfo{person}{Yuhang Li}, \bibinfo{person}{Xin Dong}, {and} \bibinfo{person}{Wei Wang}.} \bibinfo{year}{2020}\natexlab{}.
\newblock \showarticletitle{Additive Powers-of-Two Quantization: An Efficient Non-uniform Discretization for Neural Networks}. In \bibinfo{booktitle}{\emph{International Conference on Learning Representations (ICLR)}}.
\newblock
\urldef\tempurl%
\url{https://openreview.net/forum?id=BkgXT24tDS}
\showURL{%
\tempurl}


\bibitem[Li et~al\mbox{.}(2022b)]%
        {li2022q}
\bibfield{author}{\bibinfo{person}{Yanjing Li}, \bibinfo{person}{Sheng Xu}, \bibinfo{person}{Baochang Zhang}, \bibinfo{person}{Xianbin Cao}, \bibinfo{person}{Peng Gao}, {and} \bibinfo{person}{Guodong Guo}.} \bibinfo{year}{2022}\natexlab{b}.
\newblock \showarticletitle{Q-ViT: Accurate and Fully Quantized Low-bit Vision Transformer}. In \bibinfo{booktitle}{\emph{Advances in Neural Information Processing Systems (NeurIPS)}}.
\newblock
\urldef\tempurl%
\url{https://openreview.net/forum?id=fU-m9kQe0ke}
\showURL{%
\tempurl}


\bibitem[Li et~al\mbox{.}(2021c)]%
        {li2021localvit}
\bibfield{author}{\bibinfo{person}{Yawei Li}, \bibinfo{person}{Kai Zhang}, \bibinfo{person}{Jiezhang Cao}, \bibinfo{person}{Radu Timofte}, {and} \bibinfo{person}{Luc~Van Gool}.} \bibinfo{year}{2021}\natexlab{c}.
\newblock \bibinfo{title}{LocalViT: Bringing Locality to Vision Transformers}.
\newblock
\newblock
\showeprint[arxiv]{2104.05707}~[cs.CV]


\bibitem[Li et~al\mbox{.}(2022a)]%
        {liauto}
\bibfield{author}{\bibinfo{person}{Zhengang Li}, \bibinfo{person}{Mengshu Sun}, \bibinfo{person}{Alec Lu}, \bibinfo{person}{Haoyu Ma}, \bibinfo{person}{Geng Yuan}, \bibinfo{person}{Yanyue Xie}, \bibinfo{person}{Hao Tang}, \bibinfo{person}{Yanyu Li}, \bibinfo{person}{Miriam Leeser}, \bibinfo{person}{Zhangyang Wang}, \bibinfo{person}{Xue Lin}, {and} \bibinfo{person}{Zhenman Fang}.} \bibinfo{year}{2022}\natexlab{a}.
\newblock \showarticletitle{Auto-vit-acc: An fpga-aware automatic acceleration framework for vision transformer with mixed-scheme quantization}. In \bibinfo{booktitle}{\emph{2022 32nd International Conference on Field-Programmable Logic and Applications (FPL)}}. IEEE, \bibinfo{pages}{109--116}.
\newblock


\bibitem[Li et~al\mbox{.}(2021b)]%
        {li2021npas}
\bibfield{author}{\bibinfo{person}{Zhengang Li}, \bibinfo{person}{Geng Yuan}, \bibinfo{person}{Wei Niu}, \bibinfo{person}{Pu Zhao}, \bibinfo{person}{Yanyu Li}, \bibinfo{person}{Yuxuan Cai}, \bibinfo{person}{Xuan Shen}, \bibinfo{person}{Zheng Zhan}, \bibinfo{person}{Zhenglun Kong}, \bibinfo{person}{Qing Jin}, {et~al\mbox{.}}} \bibinfo{year}{2021}\natexlab{b}.
\newblock \showarticletitle{NPAS: A Compiler-aware Framework of Unified Network Pruning and Architecture Search for Beyond Real-Time Mobile Acceleration}. In \bibinfo{booktitle}{\emph{Proceedings of the IEEE/CVF Conference on Computer Vision and Pattern Recognition (CVPR)}}. \bibinfo{pages}{14255--14266}.
\newblock


\bibitem[Lin et~al\mbox{.}(2022)]%
        {lin2022fq}
\bibfield{author}{\bibinfo{person}{Yang Lin}, \bibinfo{person}{Tianyu Zhang}, \bibinfo{person}{Peiqin Sun}, \bibinfo{person}{Zheng Li}, {and} \bibinfo{person}{Shuchang Zhou}.} \bibinfo{year}{2022}\natexlab{}.
\newblock \showarticletitle{FQ-ViT: Post-Training Quantization for Fully Quantized Vision Transformer}. In \bibinfo{booktitle}{\emph{Proceedings of the Thirty-First International Joint Conference on Artificial Intelligence, {IJCAI-22}}}. \bibinfo{pages}{1173--1179}.
\newblock


\bibitem[Liu et~al\mbox{.}(2018)]%
        {liu2018progressive}
\bibfield{author}{\bibinfo{person}{Chenxi Liu}, \bibinfo{person}{Barret Zoph}, \bibinfo{person}{Maxim Neumann}, \bibinfo{person}{Jonathon Shlens}, \bibinfo{person}{Wei Hua}, \bibinfo{person}{Li-Jia Li}, \bibinfo{person}{Li Fei-Fei}, \bibinfo{person}{Alan Yuille}, \bibinfo{person}{Jonathan Huang}, {and} \bibinfo{person}{Kevin Murphy}.} \bibinfo{year}{2018}\natexlab{}.
\newblock \showarticletitle{Progressive neural architecture search}. In \bibinfo{booktitle}{\emph{Proceedings of the European conference on computer vision (ECCV)}}. \bibinfo{pages}{19--34}.
\newblock


\bibitem[Liu et~al\mbox{.}(2021a)]%
        {liu2021swin}
\bibfield{author}{\bibinfo{person}{Ze Liu}, \bibinfo{person}{Yutong Lin}, \bibinfo{person}{Yue Cao}, \bibinfo{person}{Han Hu}, \bibinfo{person}{Yixuan Wei}, \bibinfo{person}{Zheng Zhang}, \bibinfo{person}{Stephen Lin}, {and} \bibinfo{person}{Baining Guo}.} \bibinfo{year}{2021}\natexlab{a}.
\newblock \showarticletitle{Swin transformer: Hierarchical vision transformer using shifted windows}. In \bibinfo{booktitle}{\emph{Proceedings of the IEEE International Conference on Computer Vision (ICCV)}}.
\newblock


\bibitem[Liu et~al\mbox{.}(2021b)]%
        {liu2021post}
\bibfield{author}{\bibinfo{person}{Zhenhua Liu}, \bibinfo{person}{Yunhe Wang}, \bibinfo{person}{Kai Han}, \bibinfo{person}{Siwei Ma}, {and} \bibinfo{person}{Wen Gao}.} \bibinfo{year}{2021}\natexlab{b}.
\newblock \showarticletitle{Post-Training Quantization for Vision Transformer}. In \bibinfo{booktitle}{\emph{Advances in Neural Information Processing Systems (NeurIPS)}}.
\newblock


\bibitem[Loshchilov and Hutter(2019)]%
        {loshchilov2018decoupled}
\bibfield{author}{\bibinfo{person}{Ilya Loshchilov} {and} \bibinfo{person}{Frank Hutter}.} \bibinfo{year}{2019}\natexlab{}.
\newblock \showarticletitle{Decoupled Weight Decay Regularization}. In \bibinfo{booktitle}{\emph{International Conference on Learning Representations (ICLR)}}.
\newblock
\urldef\tempurl%
\url{https://openreview.net/forum?id=Bkg6RiCqY7}
\showURL{%
\tempurl}


\bibitem[Lou et~al\mbox{.}(2019)]%
        {lou2019autoq}
\bibfield{author}{\bibinfo{person}{Qian Lou}, \bibinfo{person}{Feng Guo}, \bibinfo{person}{Minje Kim}, \bibinfo{person}{Lantao Liu}, {and} \bibinfo{person}{Lei Jiang}.} \bibinfo{year}{2019}\natexlab{}.
\newblock \showarticletitle{AutoQ: Automated Kernel-Wise Neural Network Quantization}. In \bibinfo{booktitle}{\emph{International Conference on Learning Representations (ICLR)}}.
\newblock


\bibitem[Miikkulainen et~al\mbox{.}(2019)]%
        {miikkulainen2019evolving}
\bibfield{author}{\bibinfo{person}{Risto Miikkulainen}, \bibinfo{person}{Jason Liang}, \bibinfo{person}{Elliot Meyerson}, \bibinfo{person}{Aditya Rawal}, \bibinfo{person}{Dan Fink}, \bibinfo{person}{Olivier Francon}, \bibinfo{person}{Bala Raju}, \bibinfo{person}{Hormoz Shahrzad}, \bibinfo{person}{Arshak Navruzyan}, \bibinfo{person}{Nigel Duffy}, {et~al\mbox{.}}} \bibinfo{year}{2019}\natexlab{}.
\newblock \showarticletitle{Evolving deep neural networks}.
\newblock In \bibinfo{booktitle}{\emph{Artificial Intelligence in the Age of Neural Networks and Brain Computing}}. \bibinfo{publisher}{Elsevier}, \bibinfo{pages}{293--312}.
\newblock


\bibitem[Pan et~al\mbox{.}(2021)]%
        {deit_s}
\bibfield{author}{\bibinfo{person}{Bowen Pan}, \bibinfo{person}{Rameswar Panda}, \bibinfo{person}{Yifan Jiang}, \bibinfo{person}{Zhangyang Wang}, \bibinfo{person}{Rogerio Feris}, {and} \bibinfo{person}{Aude Oliva}.} \bibinfo{year}{2021}\natexlab{}.
\newblock \showarticletitle{IA-RED2: Interpretability-Aware Redundancy Reduction for Vision Transformers}.
\newblock \bibinfo{journal}{\emph{Advances in Neural Information Processing Systems (NeurIPS)}}  \bibinfo{volume}{34} (\bibinfo{year}{2021}), \bibinfo{pages}{24898--24911}.
\newblock


\bibitem[Pham et~al\mbox{.}(2018)]%
        {pham2018efficient}
\bibfield{author}{\bibinfo{person}{Hieu Pham}, \bibinfo{person}{Melody Guan}, \bibinfo{person}{Barret Zoph}, \bibinfo{person}{Quoc Le}, {and} \bibinfo{person}{Jeff Dean}.} \bibinfo{year}{2018}\natexlab{}.
\newblock \showarticletitle{Efficient neural architecture search via parameters sharing}. In \bibinfo{booktitle}{\emph{Proceedings of the International Conference on Machine Learning (ICML)}}. \bibinfo{pages}{4095--4104}.
\newblock


\bibitem[PyTorch(2024)]%
        {pytorch_profiler}
\bibfield{author}{\bibinfo{person}{PyTorch}.} \bibinfo{year}{2024}\natexlab{}.
\newblock \bibinfo{title}{PYTORCH PROFILER}.
\newblock \bibinfo{howpublished}{\url{https://pytorch.org/tutorials/recipes/recipes/profiler_recipe.html}}.
\newblock
\newblock
\shownote{Last accessed Jan. 18, 2024}.


\bibitem[Radosavovic et~al\mbox{.}(2020)]%
        {radosavovic2020designing}
\bibfield{author}{\bibinfo{person}{Ilija Radosavovic}, \bibinfo{person}{Raj~Prateek Kosaraju}, \bibinfo{person}{Ross Girshick}, \bibinfo{person}{Kaiming He}, {and} \bibinfo{person}{Piotr Doll{\'a}r}.} \bibinfo{year}{2020}\natexlab{}.
\newblock \showarticletitle{Designing network design spaces}. In \bibinfo{booktitle}{\emph{Proceedings of the IEEE/CVF Conference on Computer Vision and Pattern Recognition (CVPR)}}. \bibinfo{pages}{10428--10436}.
\newblock


\bibitem[Raghu et~al\mbox{.}(2021)]%
        {raghu2021vision}
\bibfield{author}{\bibinfo{person}{Maithra Raghu}, \bibinfo{person}{Thomas Unterthiner}, \bibinfo{person}{Simon Kornblith}, \bibinfo{person}{Chiyuan Zhang}, {and} \bibinfo{person}{Alexey Dosovitskiy}.} \bibinfo{year}{2021}\natexlab{}.
\newblock \showarticletitle{Do Vision Transformers See Like Convolutional Neural Networks?}. In \bibinfo{booktitle}{\emph{Advances in Neural Information Processing Systems (NeurIPS)}}.
\newblock


\bibitem[Rastegari et~al\mbox{.}(2016)]%
        {rastegari2016xnor}
\bibfield{author}{\bibinfo{person}{Mohammad Rastegari}, \bibinfo{person}{Vicente Ordonez}, \bibinfo{person}{Joseph Redmon}, {and} \bibinfo{person}{Ali Farhadi}.} \bibinfo{year}{2016}\natexlab{}.
\newblock \showarticletitle{Xnor-net: Imagenet classification using binary convolutional neural networks}. In \bibinfo{booktitle}{\emph{Proceedings of the European Conference on Computer Vision (ECCV)}}. \bibinfo{pages}{525--542}.
\newblock


\bibitem[Real et~al\mbox{.}(2019)]%
        {real2019regularized}
\bibfield{author}{\bibinfo{person}{Esteban Real}, \bibinfo{person}{Alok Aggarwal}, \bibinfo{person}{Yanping Huang}, {and} \bibinfo{person}{Quoc~V Le}.} \bibinfo{year}{2019}\natexlab{}.
\newblock \showarticletitle{Regularized evolution for image classifier architecture search}. In \bibinfo{booktitle}{\emph{Proceedings of the AAAI Conference on Artificial Intelligence}}, Vol.~\bibinfo{volume}{33}. \bibinfo{pages}{4780--4789}.
\newblock


\bibitem[Sahni et~al\mbox{.}(2021)]%
        {sahni2021compofa}
\bibfield{author}{\bibinfo{person}{Manas Sahni}, \bibinfo{person}{Shreya Varshini}, \bibinfo{person}{Alind Khare}, {and} \bibinfo{person}{Alexey Tumanov}.} \bibinfo{year}{2021}\natexlab{}.
\newblock \showarticletitle{CompOFA: Compound Once-For-All Networks for Faster Multi-Platform Deployment}.
\newblock \bibinfo{journal}{\emph{arXiv preprint arXiv:2104.12642}} (\bibinfo{year}{2021}).
\newblock


\bibitem[Shen et~al\mbox{.}(2020)]%
        {shen2020q}
\bibfield{author}{\bibinfo{person}{Sheng Shen}, \bibinfo{person}{Zhen Dong}, \bibinfo{person}{Jiayu Ye}, \bibinfo{person}{Linjian Ma}, \bibinfo{person}{Zhewei Yao}, \bibinfo{person}{Amir Gholami}, \bibinfo{person}{Michael~W Mahoney}, {and} \bibinfo{person}{Kurt Keutzer}.} \bibinfo{year}{2020}\natexlab{}.
\newblock \showarticletitle{Q-bert: Hessian based ultra low precision quantization of bert}. In \bibinfo{booktitle}{\emph{Proceedings of the AAAI Conference on Artificial Intelligence}}, Vol.~\bibinfo{volume}{34:05}. \bibinfo{pages}{8815--8821}.
\newblock


\bibitem[Sun et~al\mbox{.}(2022)]%
        {sun2022film}
\bibfield{author}{\bibinfo{person}{Mengshu Sun}, \bibinfo{person}{Zhengang Li}, \bibinfo{person}{Alec Lu}, \bibinfo{person}{Yanyu Li}, \bibinfo{person}{Sung-En Chang}, \bibinfo{person}{Xiaolong Ma}, \bibinfo{person}{Xue Lin}, {and} \bibinfo{person}{Zhenman Fang}.} \bibinfo{year}{2022}\natexlab{}.
\newblock \showarticletitle{Film-qnn: Efficient fpga acceleration of deep neural networks with intra-layer, mixed-precision quantization}. In \bibinfo{booktitle}{\emph{Proceedings of the 2022 ACM/SIGDA International Symposium on Field-Programmable Gate Arrays}}. \bibinfo{pages}{134--145}.
\newblock


\bibitem[Tan et~al\mbox{.}(2019)]%
        {tan2019mnasnet}
\bibfield{author}{\bibinfo{person}{Mingxing Tan}, \bibinfo{person}{Bo Chen}, \bibinfo{person}{Ruoming Pang}, \bibinfo{person}{Vijay Vasudevan}, \bibinfo{person}{Mark Sandler}, \bibinfo{person}{Andrew Howard}, {and} \bibinfo{person}{Quoc~V Le}.} \bibinfo{year}{2019}\natexlab{}.
\newblock \showarticletitle{Mnasnet: Platform-aware neural architecture search for mobile}. In \bibinfo{booktitle}{\emph{Proceedings of the IEEE/CVF Conference on Computer Vision and Pattern Recognition (CVPR)}}. \bibinfo{pages}{2820--2828}.
\newblock


\bibitem[Touvron et~al\mbox{.}(2021a)]%
        {Touvron2021TrainingDI}
\bibfield{author}{\bibinfo{person}{Hugo Touvron}, \bibinfo{person}{Matthieu Cord}, \bibinfo{person}{Matthijs Douze}, \bibinfo{person}{Francisco Massa}, \bibinfo{person}{Alexandre Sablayrolles}, {and} \bibinfo{person}{Herv{\'e} J{\'e}gou}.} \bibinfo{year}{2021}\natexlab{a}.
\newblock \showarticletitle{Training data-efficient image transformers \& distillation through attention}. In \bibinfo{booktitle}{\emph{Proceedings of the International Conference on Machine Learning (ICML)}}. PMLR, \bibinfo{pages}{10347--10357}.
\newblock


\bibitem[Touvron et~al\mbox{.}(2021b)]%
        {touvron2021going}
\bibfield{author}{\bibinfo{person}{Hugo Touvron}, \bibinfo{person}{Matthieu Cord}, \bibinfo{person}{Alexandre Sablayrolles}, \bibinfo{person}{Gabriel Synnaeve}, {and} \bibinfo{person}{Herv{\'e} J{\'e}gou}.} \bibinfo{year}{2021}\natexlab{b}.
\newblock \showarticletitle{Going deeper with image transformers}. In \bibinfo{booktitle}{\emph{Proceedings of the IEEE/CVF International Conference on Computer Vision (ICCV)}}. \bibinfo{pages}{32--42}.
\newblock


\bibitem[Touvron et~al\mbox{.}(2023)]%
        {touvron2023llama}
\bibfield{author}{\bibinfo{person}{Hugo Touvron}, \bibinfo{person}{Thibaut Lavril}, \bibinfo{person}{Gautier Izacard}, \bibinfo{person}{Xavier Martinet}, \bibinfo{person}{Marie-Anne Lachaux}, \bibinfo{person}{Timoth{\'e}e Lacroix}, \bibinfo{person}{Baptiste Rozi{\`e}re}, \bibinfo{person}{Naman Goyal}, \bibinfo{person}{Eric Hambro}, \bibinfo{person}{Faisal Azhar}, {et~al\mbox{.}}} \bibinfo{year}{2023}\natexlab{}.
\newblock \showarticletitle{Llama: Open and efficient foundation language models}.
\newblock \bibinfo{journal}{\emph{arXiv preprint arXiv:2302.13971}} (\bibinfo{year}{2023}).
\newblock


\bibitem[Uhlich et~al\mbox{.}(2020)]%
        {uhlich2019mixed}
\bibfield{author}{\bibinfo{person}{Stefan Uhlich}, \bibinfo{person}{Lukas Mauch}, \bibinfo{person}{Fabien Cardinaux}, \bibinfo{person}{Kazuki Yoshiyama}, \bibinfo{person}{Javier~Alonso Garcia}, \bibinfo{person}{Stephen Tiedemann}, \bibinfo{person}{Thomas Kemp}, {and} \bibinfo{person}{Akira Nakamura}.} \bibinfo{year}{2020}\natexlab{}.
\newblock \showarticletitle{Mixed Precision DNNs: All you need is a good parametrization}. In \bibinfo{booktitle}{\emph{International Conference on Learning Representations (ICLR)}}.
\newblock


\bibitem[Vaswani et~al\mbox{.}(2017)]%
        {vaswani2017attention}
\bibfield{author}{\bibinfo{person}{Ashish Vaswani}, \bibinfo{person}{Noam Shazeer}, \bibinfo{person}{Niki Parmar}, \bibinfo{person}{Jakob Uszkoreit}, \bibinfo{person}{Llion Jones}, \bibinfo{person}{Aidan~N Gomez}, \bibinfo{person}{{\L}ukasz Kaiser}, {and} \bibinfo{person}{Illia Polosukhin}.} \bibinfo{year}{2017}\natexlab{}.
\newblock \showarticletitle{Attention is all you need}. In \bibinfo{booktitle}{\emph{Advances in Neural Information Processing Systems (NeurIPS)}}. \bibinfo{pages}{5998--6008}.
\newblock


\bibitem[Wang et~al\mbox{.}(2021a)]%
        {wang2021attentivenas}
\bibfield{author}{\bibinfo{person}{Dilin Wang}, \bibinfo{person}{Meng Li}, \bibinfo{person}{Chengyue Gong}, {and} \bibinfo{person}{Vikas Chandra}.} \bibinfo{year}{2021}\natexlab{a}.
\newblock \showarticletitle{Attentivenas: Improving neural architecture search via attentive sampling}. In \bibinfo{booktitle}{\emph{Proceedings of the IEEE/CVF Conference on Computer Vision and Pattern Recognition (CVPR)}}. \bibinfo{pages}{6418--6427}.
\newblock


\bibitem[Wang et~al\mbox{.}(2020b)]%
        {wang2020hat}
\bibfield{author}{\bibinfo{person}{Hanrui Wang}, \bibinfo{person}{Zhanghao Wu}, \bibinfo{person}{Zhijian Liu}, \bibinfo{person}{Han Cai}, \bibinfo{person}{Ligeng Zhu}, \bibinfo{person}{Chuang Gan}, {and} \bibinfo{person}{Song Han}.} \bibinfo{year}{2020}\natexlab{b}.
\newblock \showarticletitle{Hat: Hardware-aware transformers for efficient natural language processing}.
\newblock \bibinfo{journal}{\emph{arXiv:2005.14187}} (\bibinfo{year}{2020}).
\newblock


\bibitem[Wang et~al\mbox{.}(2019)]%
        {wang2019haq}
\bibfield{author}{\bibinfo{person}{Kuan Wang}, \bibinfo{person}{Zhijian Liu}, \bibinfo{person}{Yujun Lin}, \bibinfo{person}{Ji Lin}, {and} \bibinfo{person}{Song Han}.} \bibinfo{year}{2019}\natexlab{}.
\newblock \showarticletitle{HAQ: Hardware-Aware Automated Quantization with Mixed Precision}. In \bibinfo{booktitle}{\emph{Proceedings of the IEEE/CVF Conference on Computer Vision and Pattern Recognition (CVPR)}}.
\newblock


\bibitem[Wang et~al\mbox{.}(2020a)]%
        {wang2020apq}
\bibfield{author}{\bibinfo{person}{Tianzhe Wang}, \bibinfo{person}{Kuan Wang}, \bibinfo{person}{Han Cai}, \bibinfo{person}{Ji Lin}, \bibinfo{person}{Zhijian Liu}, \bibinfo{person}{Hanrui Wang}, \bibinfo{person}{Yujun Lin}, {and} \bibinfo{person}{Song Han}.} \bibinfo{year}{2020}\natexlab{a}.
\newblock \showarticletitle{Apq: Joint search for network architecture, pruning and quantization policy}. In \bibinfo{booktitle}{\emph{Proceedings of the IEEE/CVF Conference on Computer Vision and Pattern Recognition}}. \bibinfo{pages}{2078--2087}.
\newblock


\bibitem[Wang et~al\mbox{.}(2021b)]%
        {wang2021pyramid}
\bibfield{author}{\bibinfo{person}{Wenhai Wang}, \bibinfo{person}{Enze Xie}, \bibinfo{person}{Xiang Li}, \bibinfo{person}{Deng-Ping Fan}, \bibinfo{person}{Kaitao Song}, \bibinfo{person}{Ding Liang}, \bibinfo{person}{Tong Lu}, \bibinfo{person}{Ping Luo}, {and} \bibinfo{person}{Ling Shao}.} \bibinfo{year}{2021}\natexlab{b}.
\newblock \showarticletitle{Pyramid Vision Transformer: A Versatile Backbone for Dense Prediction without Convolutions}. In \bibinfo{booktitle}{\emph{Proceedings of the IEEE/CVF International Conference on Computer Vision (ICCV)}}.
\newblock


\bibitem[Wang et~al\mbox{.}(2022)]%
        {wang2022QGTC}
\bibfield{author}{\bibinfo{person}{Yuke Wang}, \bibinfo{person}{Boyuan Feng}, {and} \bibinfo{person}{Yufei Ding}.} \bibinfo{year}{2022}\natexlab{}.
\newblock \showarticletitle{QGTC: accelerating quantized graph neural networks via GPU tensor core}. In \bibinfo{booktitle}{\emph{Proceedings of the 27th ACM SIGPLAN Symposium on Principles and Practice of Parallel Programming}}. \bibinfo{pages}{107–119}.
\newblock


\bibitem[Wu et~al\mbox{.}(2019)]%
        {wu2019fbnet}
\bibfield{author}{\bibinfo{person}{Bichen Wu}, \bibinfo{person}{Xiaoliang Dai}, \bibinfo{person}{Peizhao Zhang}, \bibinfo{person}{Yanghan Wang}, \bibinfo{person}{Fei Sun}, \bibinfo{person}{Yiming Wu}, \bibinfo{person}{Yuandong Tian}, \bibinfo{person}{Peter Vajda}, \bibinfo{person}{Yangqing Jia}, {and} \bibinfo{person}{Kurt Keutzer}.} \bibinfo{year}{2019}\natexlab{}.
\newblock \showarticletitle{Fbnet: Hardware-aware efficient convnet design via differentiable neural architecture search}. In \bibinfo{booktitle}{\emph{Proceedings of the IEEE/CVF Conference on Computer Vision and Pattern Recognition (CVPR)}}. \bibinfo{pages}{10734--10742}.
\newblock


\bibitem[Wu et~al\mbox{.}(2018)]%
        {wu2018mixed}
\bibfield{author}{\bibinfo{person}{Bichen Wu}, \bibinfo{person}{Yanghan Wang}, \bibinfo{person}{Peizhao Zhang}, \bibinfo{person}{Yuandong Tian}, \bibinfo{person}{Peter Vajda}, {and} \bibinfo{person}{Kurt Keutzer}.} \bibinfo{year}{2018}\natexlab{}.
\newblock \showarticletitle{Mixed precision quantization of convnets via differentiable neural architecture search}.
\newblock \bibinfo{journal}{\emph{arXiv:1812.00090}} (\bibinfo{year}{2018}).
\newblock


\bibitem[Wu et~al\mbox{.}(2021)]%
        {wu2021cvt}
\bibfield{author}{\bibinfo{person}{Haiping Wu}, \bibinfo{person}{Bin Xiao}, \bibinfo{person}{Noel Codella}, \bibinfo{person}{Mengchen Liu}, \bibinfo{person}{Xiyang Dai}, \bibinfo{person}{Lu Yuan}, {and} \bibinfo{person}{Lei Zhang}.} \bibinfo{year}{2021}\natexlab{}.
\newblock \showarticletitle{Cvt: Introducing convolutions to vision transformers}.
\newblock \bibinfo{journal}{\emph{arXiv preprint arXiv:2103.15808}} (\bibinfo{year}{2021}).
\newblock


\bibitem[Xie et~al\mbox{.}(2017)]%
        {xie2017aggregated}
\bibfield{author}{\bibinfo{person}{Saining Xie}, \bibinfo{person}{Ross Girshick}, \bibinfo{person}{Piotr Doll{\'a}r}, \bibinfo{person}{Zhuowen Tu}, {and} \bibinfo{person}{Kaiming He}.} \bibinfo{year}{2017}\natexlab{}.
\newblock \showarticletitle{Aggregated residual transformations for deep neural networks}. In \bibinfo{booktitle}{\emph{Proceedings of the IEEE/CVF Conference on Computer Vision and Pattern Recognition (CVPR)}}.
\newblock


\bibitem[Xilinx(2017)]%
        {Xilinx-dspPack-Int8}
\bibfield{author}{\bibinfo{person}{Xilinx}.} \bibinfo{year}{2017}\natexlab{}.
\newblock \bibinfo{title}{Deep Learning with INT8 Optimization on Xilinx Devices}.
\newblock \bibinfo{howpublished}{\url{https://docs.xilinx.com/v/u/en-US/wp486-deep-learning-int8}}.
\newblock
\newblock
\shownote{Last accessed Mar. 28, 2022}.


\bibitem[Xilinx(2020a)]%
        {Xilinx-dspPack-Int4}
\bibfield{author}{\bibinfo{person}{Xilinx}.} \bibinfo{year}{2020}\natexlab{a}.
\newblock \bibinfo{title}{Convolutional Neural Network with INT4 Optimization on Xilinx Devices}.
\newblock \bibinfo{howpublished}{\url{https://docs.xilinx.com/v/u/en-US/wp521-4bit-optimization}}.
\newblock
\newblock
\shownote{Last accessed Mar. 28, 2022}.


\bibitem[Xilinx(2020b)]%
        {url-vivado}
\bibfield{author}{\bibinfo{person}{Xilinx}.} \bibinfo{year}{2020}\natexlab{b}.
\newblock \bibinfo{title}{{Vivado Design Suite}}.
\newblock \bibinfo{howpublished}{\url{https://www.xilinx.com/products/design-tools/vivado.html}}.
\newblock
\newblock
\shownote{Last accessed Aug. 28, 2022}.


\bibitem[You et~al\mbox{.}(2020)]%
        {you2020greedynas}
\bibfield{author}{\bibinfo{person}{Shan You}, \bibinfo{person}{Tao Huang}, \bibinfo{person}{Mingmin Yang}, \bibinfo{person}{Fei Wang}, \bibinfo{person}{Chen Qian}, {and} \bibinfo{person}{Changshui Zhang}.} \bibinfo{year}{2020}\natexlab{}.
\newblock \showarticletitle{GreedyNAS: Towards Fast One-Shot NAS with Greedy Supernet}. In \bibinfo{booktitle}{\emph{Proceedings of the IEEE/CVF Conference on Computer Vision and Pattern Recognition (CVPR)}}. \bibinfo{pages}{1999--2008}.
\newblock


\bibitem[Yu et~al\mbox{.}(2020)]%
        {yu2020bignas}
\bibfield{author}{\bibinfo{person}{Jiahui Yu}, \bibinfo{person}{Pengchong Jin}, \bibinfo{person}{Hanxiao Liu}, \bibinfo{person}{Gabriel Bender}, \bibinfo{person}{Pieter-Jan Kindermans}, \bibinfo{person}{Mingxing Tan}, \bibinfo{person}{Thomas Huang}, \bibinfo{person}{Xiaodan Song}, \bibinfo{person}{Ruoming Pang}, {and} \bibinfo{person}{Quoc Le}.} \bibinfo{year}{2020}\natexlab{}.
\newblock \showarticletitle{Bignas: Scaling up neural architecture search with big single-stage models}. In \bibinfo{booktitle}{\emph{Proceedings of the European Conference on Computer Vision (ECCV)}}. \bibinfo{pages}{702--717}.
\newblock


\bibitem[Yuan et~al\mbox{.}(2021a)]%
        {yuan2021tokens}
\bibfield{author}{\bibinfo{person}{Li Yuan}, \bibinfo{person}{Yunpeng Chen}, \bibinfo{person}{Tao Wang}, \bibinfo{person}{Weihao Yu}, \bibinfo{person}{Yujun Shi}, \bibinfo{person}{Zi-Hang Jiang}, \bibinfo{person}{Francis~E.H. Tay}, \bibinfo{person}{Jiashi Feng}, {and} \bibinfo{person}{Shuicheng Yan}.} \bibinfo{year}{2021}\natexlab{a}.
\newblock \showarticletitle{Tokens-to-Token ViT: Training Vision Transformers From Scratch on ImageNet}. In \bibinfo{booktitle}{\emph{Proceedings of the IEEE/CVF International Conference on Computer Vision (ICCV)}}. \bibinfo{pages}{558--567}.
\newblock


\bibitem[Yuan et~al\mbox{.}(2021b)]%
        {Yuan_2021_ICCV}
\bibfield{author}{\bibinfo{person}{Li Yuan}, \bibinfo{person}{Yunpeng Chen}, \bibinfo{person}{Tao Wang}, \bibinfo{person}{Weihao Yu}, \bibinfo{person}{Yujun Shi}, \bibinfo{person}{Zi-Hang Jiang}, \bibinfo{person}{Francis~E.H. Tay}, \bibinfo{person}{Jiashi Feng}, {and} \bibinfo{person}{Shuicheng Yan}.} \bibinfo{year}{2021}\natexlab{b}.
\newblock \showarticletitle{Tokens-to-Token ViT: Training Vision Transformers From Scratch on ImageNet}. In \bibinfo{booktitle}{\emph{Proceedings of the IEEE/CVF International Conference on Computer Vision (ICCV)}}. \bibinfo{pages}{558--567}.
\newblock


\bibitem[Zafrir et~al\mbox{.}(2019)]%
        {zafrir2019q8bert}
\bibfield{author}{\bibinfo{person}{Ofir Zafrir}, \bibinfo{person}{Guy Boudoukh}, \bibinfo{person}{Peter Izsak}, {and} \bibinfo{person}{Moshe Wasserblat}.} \bibinfo{year}{2019}\natexlab{}.
\newblock \showarticletitle{Q8bert: Quantized 8bit bert}. In \bibinfo{booktitle}{\emph{2019 Fifth Workshop on Energy Efficient Machine Learning and Cognitive Computing-NeurIPS Edition (EMC2-NIPS)}}. \bibinfo{pages}{36--39}.
\newblock


\bibitem[Zhang et~al\mbox{.}(2020)]%
        {zhang2020ternarybert}
\bibfield{author}{\bibinfo{person}{Wei Zhang}, \bibinfo{person}{Lu Hou}, \bibinfo{person}{Yichun Yin}, \bibinfo{person}{Lifeng Shang}, \bibinfo{person}{Xiao Chen}, \bibinfo{person}{Xin Jiang}, {and} \bibinfo{person}{Qun Liu}.} \bibinfo{year}{2020}\natexlab{}.
\newblock \showarticletitle{Ternarybert: Distillation-aware ultra-low bit bert}. In \bibinfo{booktitle}{\emph{Proceedings of the 2020 Conference on Empirical Methods in Natural Language Processing (EMNLP)}}.
\newblock


\bibitem[Zhao et~al\mbox{.}(2021)]%
        {zhao2021few}
\bibfield{author}{\bibinfo{person}{Yiyang Zhao}, \bibinfo{person}{Linnan Wang}, \bibinfo{person}{Yuandong Tian}, \bibinfo{person}{Rodrigo Fonseca}, {and} \bibinfo{person}{Tian Guo}.} \bibinfo{year}{2021}\natexlab{}.
\newblock \showarticletitle{Few-shot neural architecture search}. In \bibinfo{booktitle}{\emph{Proceedings of the International Conference on Machine Learning (ICML)}}. PMLR, \bibinfo{pages}{12707--12718}.
\newblock


\bibitem[Zhong et~al\mbox{.}(2018)]%
        {zhong2018practical}
\bibfield{author}{\bibinfo{person}{Zhao Zhong}, \bibinfo{person}{Junjie Yan}, \bibinfo{person}{Wei Wu}, \bibinfo{person}{Jing Shao}, {and} \bibinfo{person}{Cheng-Lin Liu}.} \bibinfo{year}{2018}\natexlab{}.
\newblock \showarticletitle{Practical block-wise neural network architecture generation}. In \bibinfo{booktitle}{\emph{Proceedings of the IEEE/CVF Conference on Computer Vision and Pattern Recognition (CVPR)}}. \bibinfo{pages}{2423--2432}.
\newblock


\bibitem[Zhou et~al\mbox{.}(2016)]%
        {zhou2016dorefa}
\bibfield{author}{\bibinfo{person}{Shuchang Zhou}, \bibinfo{person}{Yuxin Wu}, \bibinfo{person}{Zekun Ni}, \bibinfo{person}{Xinyu Zhou}, \bibinfo{person}{He Wen}, {and} \bibinfo{person}{Yuheng Zou}.} \bibinfo{year}{2016}\natexlab{}.
\newblock \showarticletitle{Dorefa-net: Training low bitwidth convolutional neural networks with low bitwidth gradients}.
\newblock \bibinfo{journal}{\emph{arXiv:1606.06160}} (\bibinfo{year}{2016}).
\newblock


\bibitem[Zoph and Le(2017)]%
        {zoph2016neural}
\bibfield{author}{\bibinfo{person}{Barret Zoph} {and} \bibinfo{person}{Quoc~V. Le}.} \bibinfo{year}{2017}\natexlab{}.
\newblock \showarticletitle{Neural Architecture Search with Reinforcement Learning}. In \bibinfo{booktitle}{\emph{International Conference on Learning Representations (ICLR)}}.
\newblock


\bibitem[Zoph et~al\mbox{.}(2018)]%
        {zoph2018learning}
\bibfield{author}{\bibinfo{person}{Barret Zoph}, \bibinfo{person}{Vijay Vasudevan}, \bibinfo{person}{Jonathon Shlens}, {and} \bibinfo{person}{Quoc~V Le}.} \bibinfo{year}{2018}\natexlab{}.
\newblock \showarticletitle{Learning transferable architectures for scalable image recognition}. In \bibinfo{booktitle}{\emph{Proceedings of the IEEE/CVF Conference on Computer Vision and Pattern Recognition (CVPR)}}. \bibinfo{pages}{8697--8710}.
\newblock


\end{thebibliography}

%%
%% If your work has an appendix, this is the place to put it.

\end{document}